\documentclass{article}
\usepackage{spconf,amsmath,graphicx}
\pdfoutput = 1

\title{Towards CT-quality ultrasound imaging using deep learning}

 \name{Sanketh Vedula$^{\star,\dagger}$\thanks{$^{\star}$ equal contribution} Ortal Senouf $^{\star,\dagger}$  Alex M. Bronstein$^{\dagger}$ Oleg V. Michailovich$^{\ddagger}$ Michael Zibulevsky$^{\dagger}$ }

 \address{$^{\dagger}$ Technion -- Israel Institute of Technology \\
     $^{\ddagger}$ Electrical and Computer Engineering, University of Waterloo, Canada\\
}

\begin{document}
%
\maketitle
\begin{abstract}
The cost-effectiveness and practical harmlessness of ultrasound imaging have made it one of the most widespread tools for medical diagnosis. Unfortunately, the beam-forming based image formation produces granular speckle noise, blurring, shading and other artifacts. To overcome these effects, the ultimate goal would be to reconstruct the tissue acoustic properties by solving a full wave propagation inverse problem. In this work, we make a step towards this goal, using Multi-Resolution Convolutional Neural Networks (CNN). As a result, we are able to reconstruct CT-quality images from the reflected ultrasound radio-frequency(RF) data obtained by simulation from real CT scans of a human body.  We also show that CNN is able to imitate existing computationally heavy despeckling methods, thereby saving orders of magnitude in computations and making them amenable to real-time applications.
\end{abstract}
%
\section{Introduction}\label{sec:intro}
Medical ultrasound (US) is a widespread imaging modality owing its popularity to cost efficiency, portability, speed, and lack of harmful ionizing radiation. Unfortunately, the quality of medical ultrasound scans is typically affected by a number of artifacts, some of which are rooted in the physics of ultrasound image formation. Thus, for example, the band-limited characteristics of acoustic beamforming are generally responsible for the reduced spatial resolution. Additionally, the coherent nature of ultrasound image formation makes the resulting scans suffer from the degrading effect of speckle noise, which tends to reduce image contrast and obscure diagnostically relevant details. As a result, ultrasound imaging has limited applicability to clinical tasks relying on tissue characterization.

The problem of speckle noise contamination is traditionally solved by means of {\it image despeckling}. Thus, in the past decade, the arsenal of image despeckling algorithms has been enhanced by particularly effective methods based on homomorphic de-noising \cite{michailovich2006despeckling}, non-local means (NLM) \cite{coupe2009nonlocal}, and BM3D filtering \cite{gan2015bm3d}. Although some of these methods have already been integrated in high-end ultrasound scanners, their elevated computational requirement and relatively long run-times prevent their deployment on portable devices or in real-time applications.

Recently, convolutional neural networks (CNN) have been extensively used for solving various problems of image processing \cite{zhang2017beyond,dong2014learning}, including some important medical imaging applications, among which are X-ray CT image reconstruction \cite{jin2017deep} and denoising \cite{yang2017ct}. In this paper, we demonstrate the applicability of CNN as a means for fast and accurate approximation of the image restoration results produced by advanced (yet, relatively slow) despeckling algorithms. We believe the availability of such solutions opens the doors to the possibility of using cutting-edge signal processing on low-cost (portable) devices and in real-time applications. In addition, we also demonstrate the applicability of CNN to the reconstruction of ``CT-quality'' ultrasound images from standard ultrasonograms.

The contributions of this paper are three-fold: Firstly, we propose an end-to-end CNN architecture for real-time despeckling of ultrasound images. 
Secondly, we show that our network successfully approximates the state-of-the-art de-speckling algorithms, and works about $\times 4$ faster on GPU, and $\times 8-260$ faster on CPU, which makes it amenable to real-time applications.
Finally, aiming to go beyond the quality of despeckling, we propose a method for training CNN for the purpose of converting regular US scans into corresponding ``CT-like'' images which feature substantially improved resolution and contrast, while preserving all the relevant anatomical and pathological information.



		
		

The remainder of the paper is organized as follows: We briefly overview the proposed methodology in Section \ref{sec:methods} and provide its experimental assessment in Section \ref{sec:majhead}. Section \ref{sec:conclusions} concludes the paper.

\section{Methods}\label{sec:methods}
\subsection{Restoration of medical ultrasound scans}\label{despeckle}
The quintessential goal of ultrasound image despeckling is to reject the spurious variations of image intensity caused by speckle noise, while preserving the anatomical integrity and discernibility of diagnostical details. More often than not, the methods of image despeckling take advantage of advanced denoising methodologies, properly adapted to cope with the peculiar statistical nature of speckle noise. In particular, the nowadays standard methods of image denoising based on multi-resolution analysis and sparse representations, Bayesian estimation and random Markov fields, diffusion-based filtering, and non-local means, have all been attempted in application to image despeckling\cite{Gupta2016Review}. In view of this fact, it is hardly surprising that image despeckling faces the same challenges as the more general methods of image denoising, chief among which is the fact that their effectiveness tends to increase {\it pro rata} with computational complexity. As a result, despite the recent advances in computational technologies, many powerful despeckling and denoising tools are still unavailable for real-time applications. The most prominent examples of such methods include NLM \cite{coupe2009nonlocal} and BM3D \cite{gan2015bm3d} filters, both of which have polynomial time complexity, even under approximative execution. Accordingly, the principal goal of this study has been to develop CNN capable of accurately approximating the effect of the above filters at much smaller run-times. 

In this work, all the proposed and reference methods are applied within the homomorphic filtering framework, which considers speckle noise to be multiplicative in its nature. Thus, a typical homomorphic filter amounts to denoising of logarithmically transformed data, followed by exponentiation of the result thus obtained. Moreover, as demonstrated in \cite{michailovich2006despeckling}, this type of despeckling becomes particularly effective if applied to the envelops of {\it deconvolved} RF data, with an addition of the procedure of outlier shrinkage. While the deconvolution is helpful in substantially decorrelating the speckle noise, the outlier shrinkage has an effect of gaussianization of noise behaviour via  suppressing the ``heavy tails'' of the noise distribution produced by the log transform. The net effect of the above preprocessing is a considerable simplification of the statistical properties of the noise that can now be treated as being approximately white and Gaussian. Consequently, in what follows, all the tested method are applied to the preprocessed data, with particular denoising methods being total-variation (TV), NLM, and BM3D filtering.   

\subsection{Data Sources}\label{ssec:data}
In order to conduct numerical experiments under controllable conditions, {\it in silico} ultrasound datasets were used. Unfortunately, in view of the fact that training of CNN normally requires tens of thousands of images, generating such data by means of standard simulation software packages (such as, e.g., k-Wave or Field-II) was found to be infeasible. As a result, the {\it in silico} data were simulated using the fast simulation scheme of \cite{Kutter2009Visualization}, which allows one to properly model all the major US artifacts, including acoustic shadowing, dispersive attenuation, refraction-related effects, etc. The simulation parameters were set so as to mimic the characteristics of a standard linear array probe with a central frequency of 5 MHz and an approximate Q-factor of 0.5. The distributions of tissue density and acoustic velocities have been deduced from the corresponding Hounsfield units of abdominal X-ray CT scans.

\subsection{CNN-based approximation}
\label{ssec:NetworkArchitecture}

In our work, we use the in-phase/quadrature (IQ) images that are derived from radio-frequency (RF) images through the process of frequency demodulation. We propose training a CNN over the dataset of pairs of IQ images and resulted conventional despeckled images to achieve a fast approximation of the despeckling algorithms mentioned in Section \ref{despeckle}. Going beyond despeckling, we aim to estimate the CT image given an ultrasound one by exploiting the fact that intensity in CT images (measured in Hounsfield units) is proportional to the tissue density. 

Similarly, ultrasound imaging depends on the speed of sound and acoustic impedance, which in turn are also directly influenced by the density of the tissue. Hence, it is clear that the X-ray CT and ultrasound imaging are affected by a common physical property of the tissue. Exploiting this fact, we propose to use CNN to approximate a ``CT-quality'' image of the tissue given its corresponding ultrasound image.

Experiments presented in this paper were conducted using a multi-resolution fully convolutional neural network (CNN) consisting of a down-sampling track followed by an up-sampling track. Our network consists of $10$ convolutional layers with symmetric skip connections from each layer in the down-sampling track to its corresponding layer in the up-sampling track, similar to \cite{Mao2016SymmetricSkip}. All the kernel sizes of convolutions have been set to be $3\times3$. The non-linearities are set to {ReLU} and training is performed on mini-batches of $6$ images using Adam optimizer, with a learning rate of $1\times10^{-4}$.

\section{Experiments}
\label{sec:majhead}
\subsection{Fast approximation of despeckling} 
\label{ssec:FFA}
\textbf{Dataset.} The US RF data (IQ images) was obtained by applying our simulator described in Section \ref{ssec:data} to CT image patches extracted from the Cancer Imaging Archive (TCIA) \cite{TCIA}. Conventional despeckling algorithms (TV, BM3D, NLM) were applied on the US IQ data and the corresponding despeckled patches were generated.\\
\\
\textbf{Training.} The dataset comprised of 8200 samples of US IQ images and the corresponding despeckled images using the conventional despeckling algorithms mentioned above. The network architecture as detailed in Section \ref{ssec:NetworkArchitecture} was implemented with the two input channels corresponding to real and imaginary parts of the complex-valued IQ image, and the output as the corresponding despeckled image. Training was performed for $50K$-$60K$ mini-batch iterations for $2$ hours. \\
\\
\textbf{Results.} Post training, 420 pairs of US IQ images and their respective despeckled patches were considered for evaluation. The performance of trained network was compared to that of the conventional despeckling algorithms. The benchmarking was performed on both Intel Xeon E5 CPU and an NVIDIA Titan X GPU. It has to be noted that the GPU implementations of BM3D and NLM methods was not available for comparison. 

\begin{figure}[t]
\makebox[0pt][l]{%
\begin{minipage}[]{\linewidth}
	\begin{tabular}{ c@{\hskip 0.001\textwidth}c@{\hskip 0.001\textwidth}c@{\hskip 0.001\textwidth}c@{\hskip 0.001\textwidth}c} 

		\includegraphics[width = 0.24\textwidth]{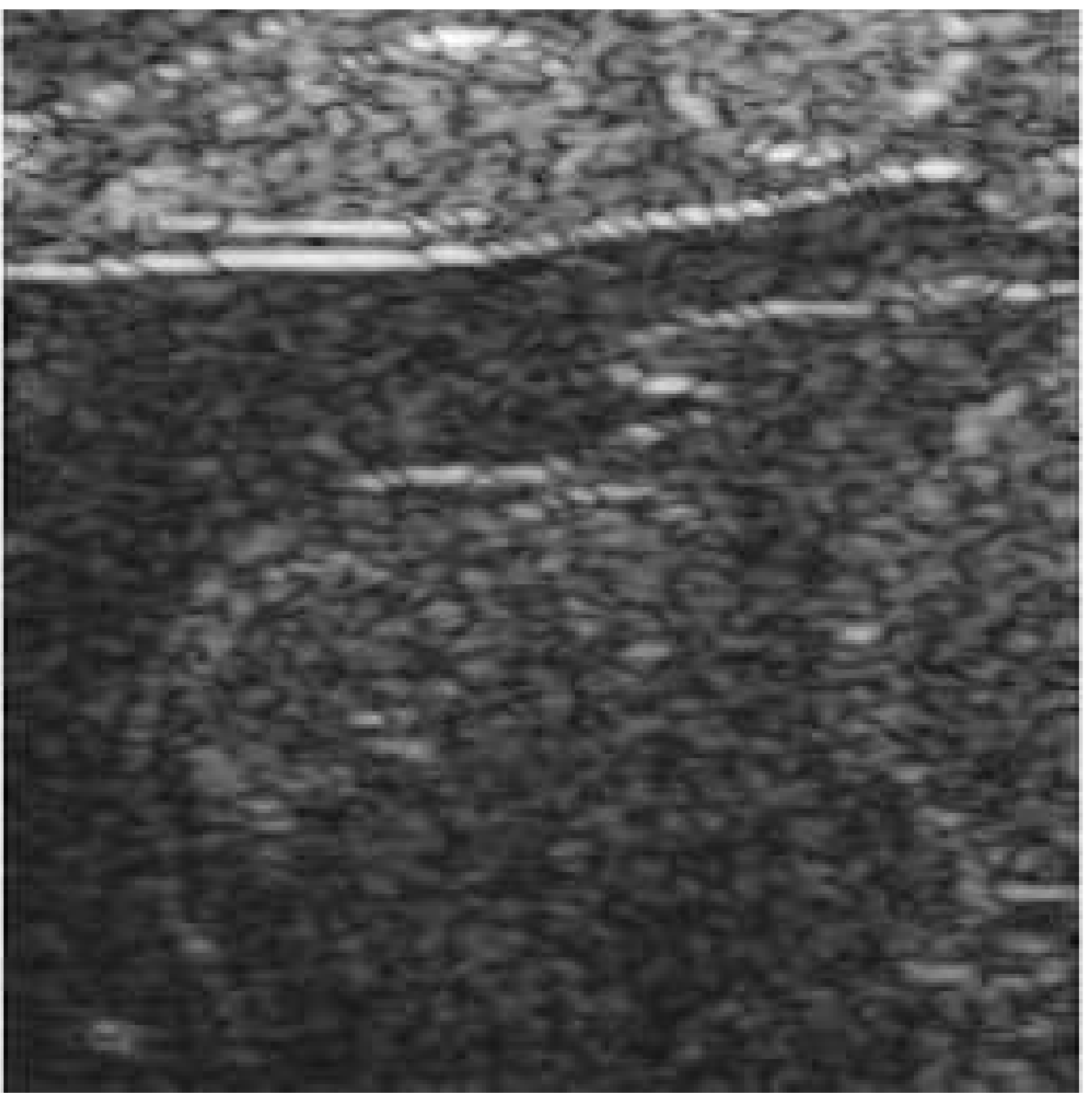} &
		\includegraphics[width = 0.24\textwidth]{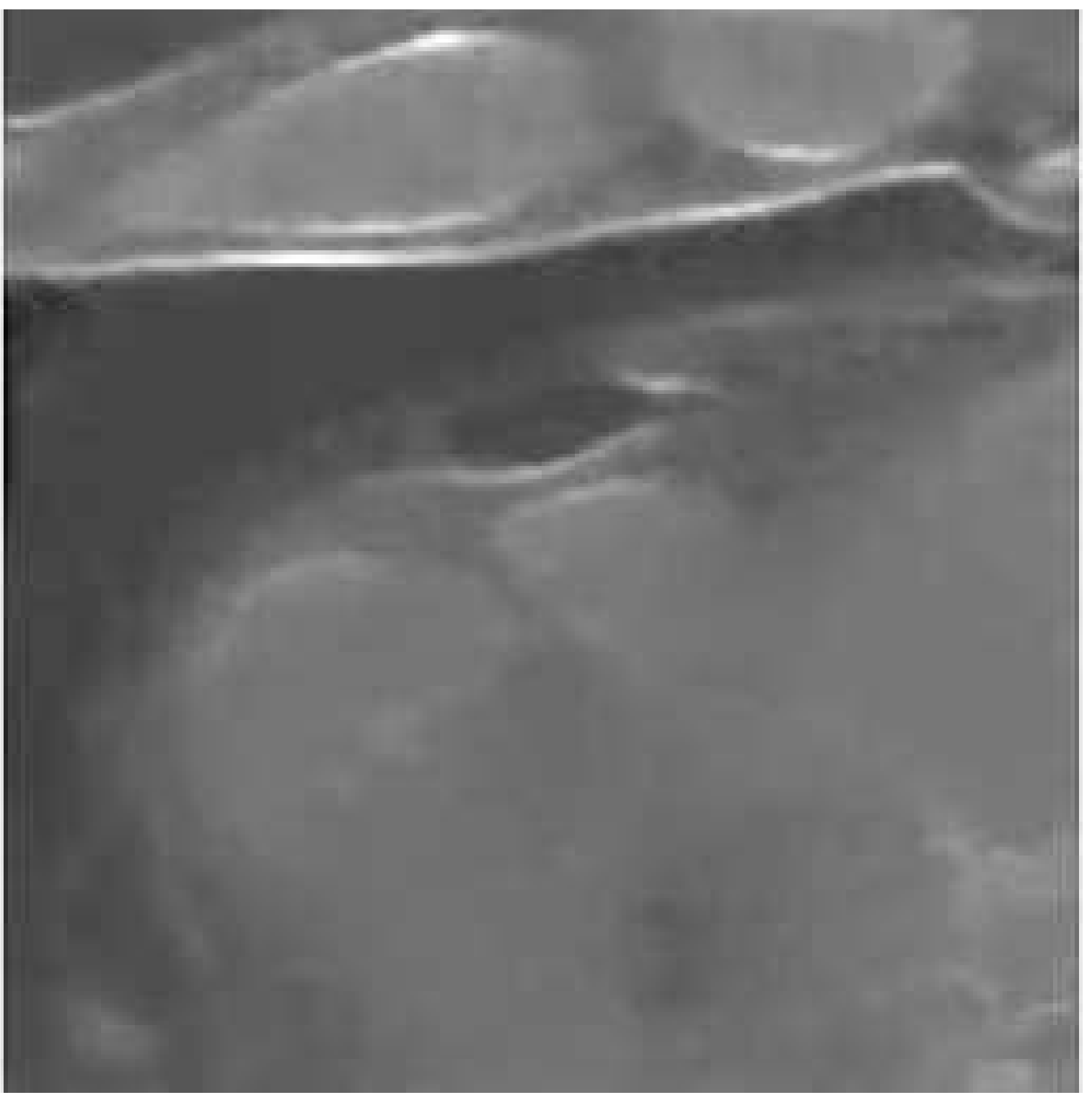} &
		\includegraphics[width = 0.24\textwidth]{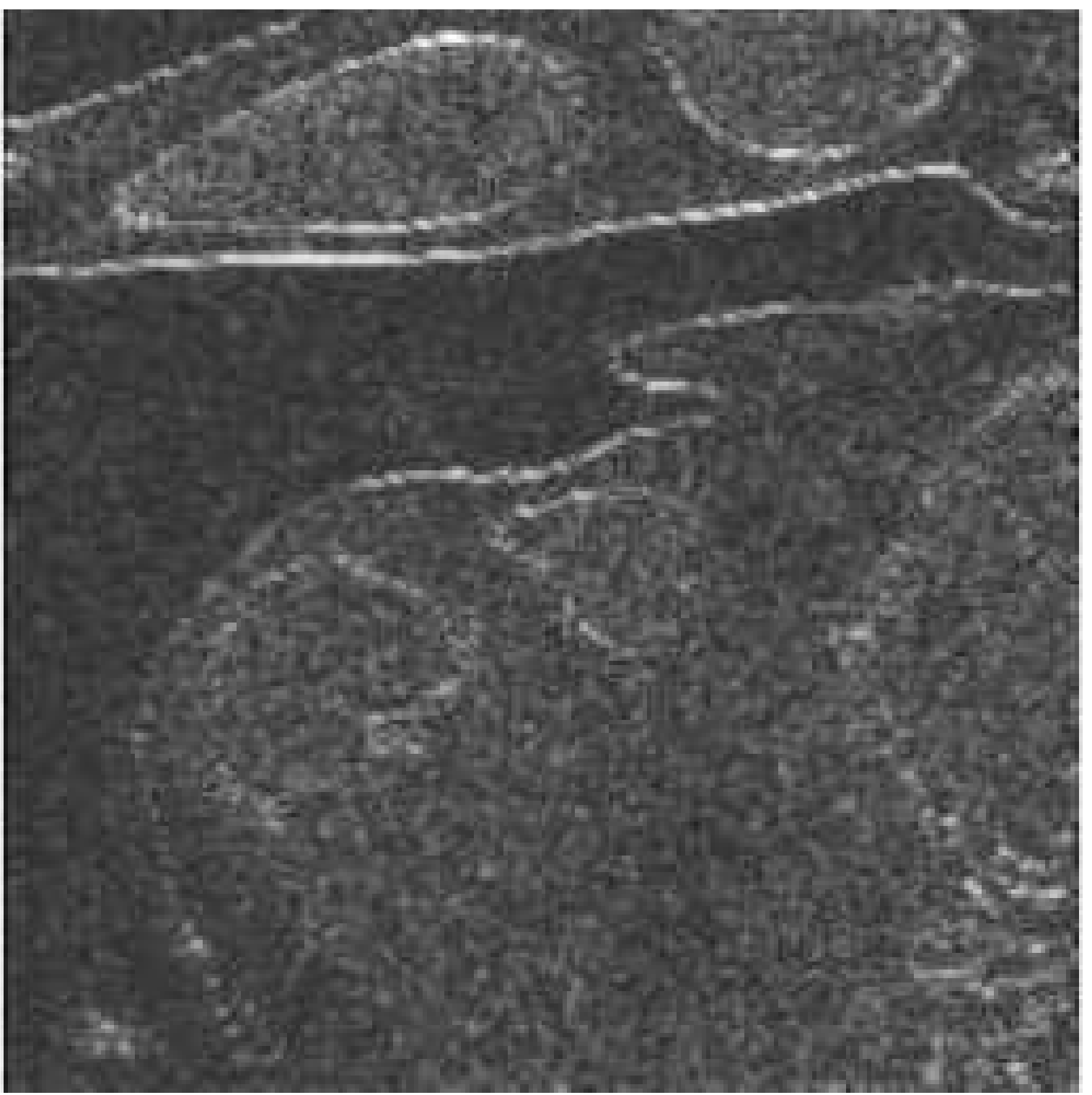} &
		\includegraphics[width = 0.24\textwidth]{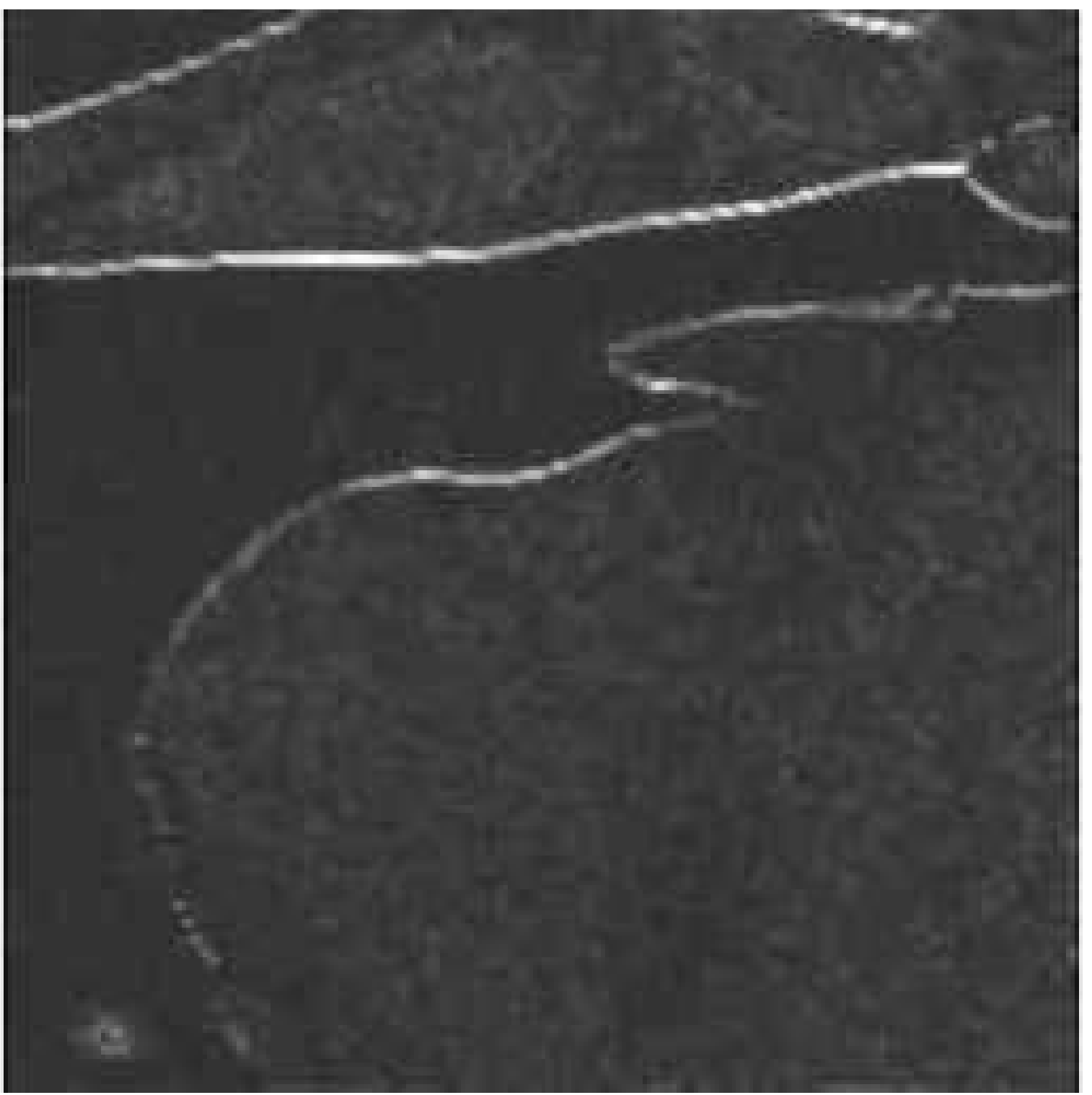} & \\  
		\small US image(a) & \small TV & \small NLM & \small BM3D \\
		 &
		\includegraphics[width = 0.24\textwidth]{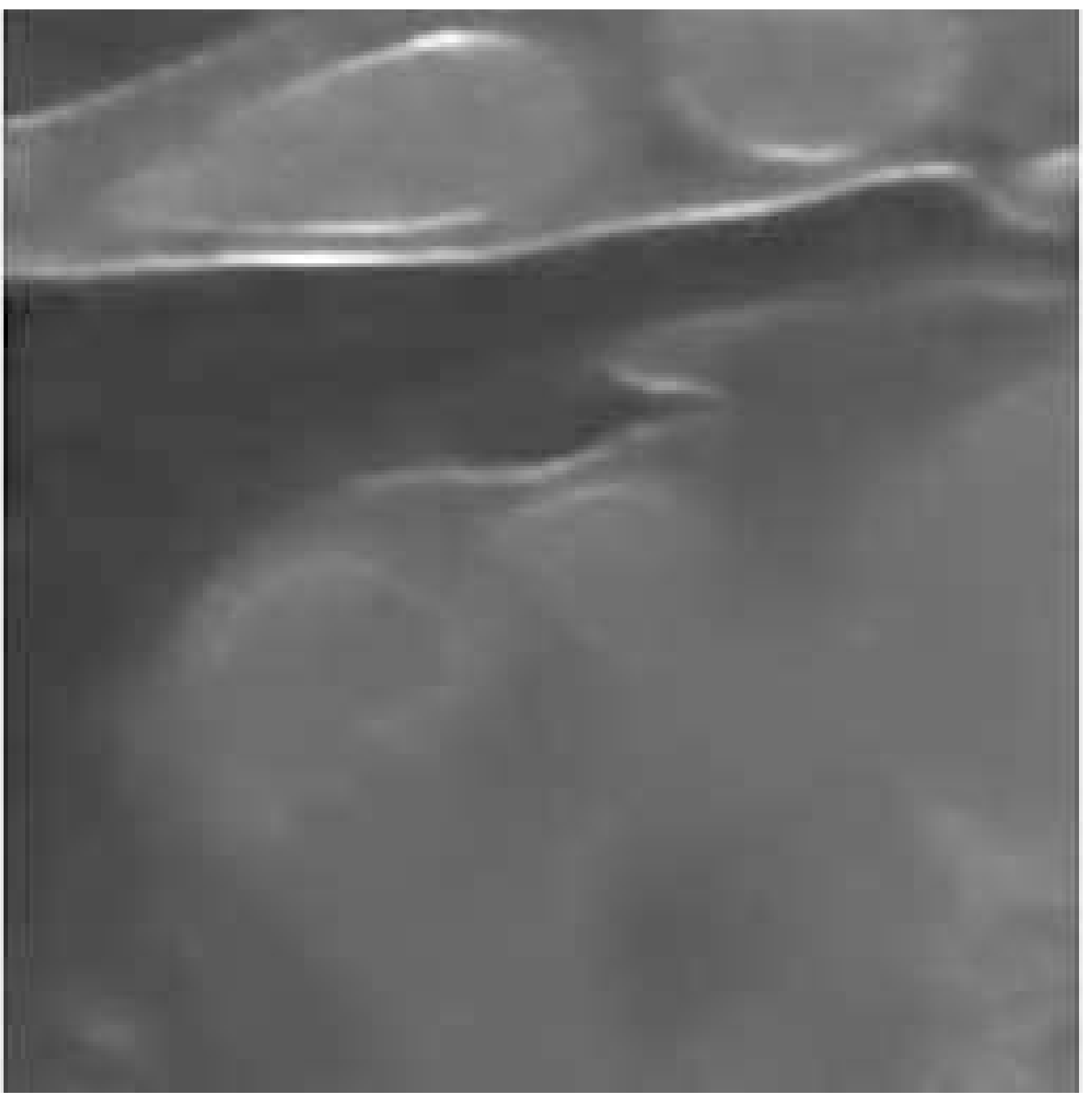} &
		\includegraphics[width = 0.24\textwidth]{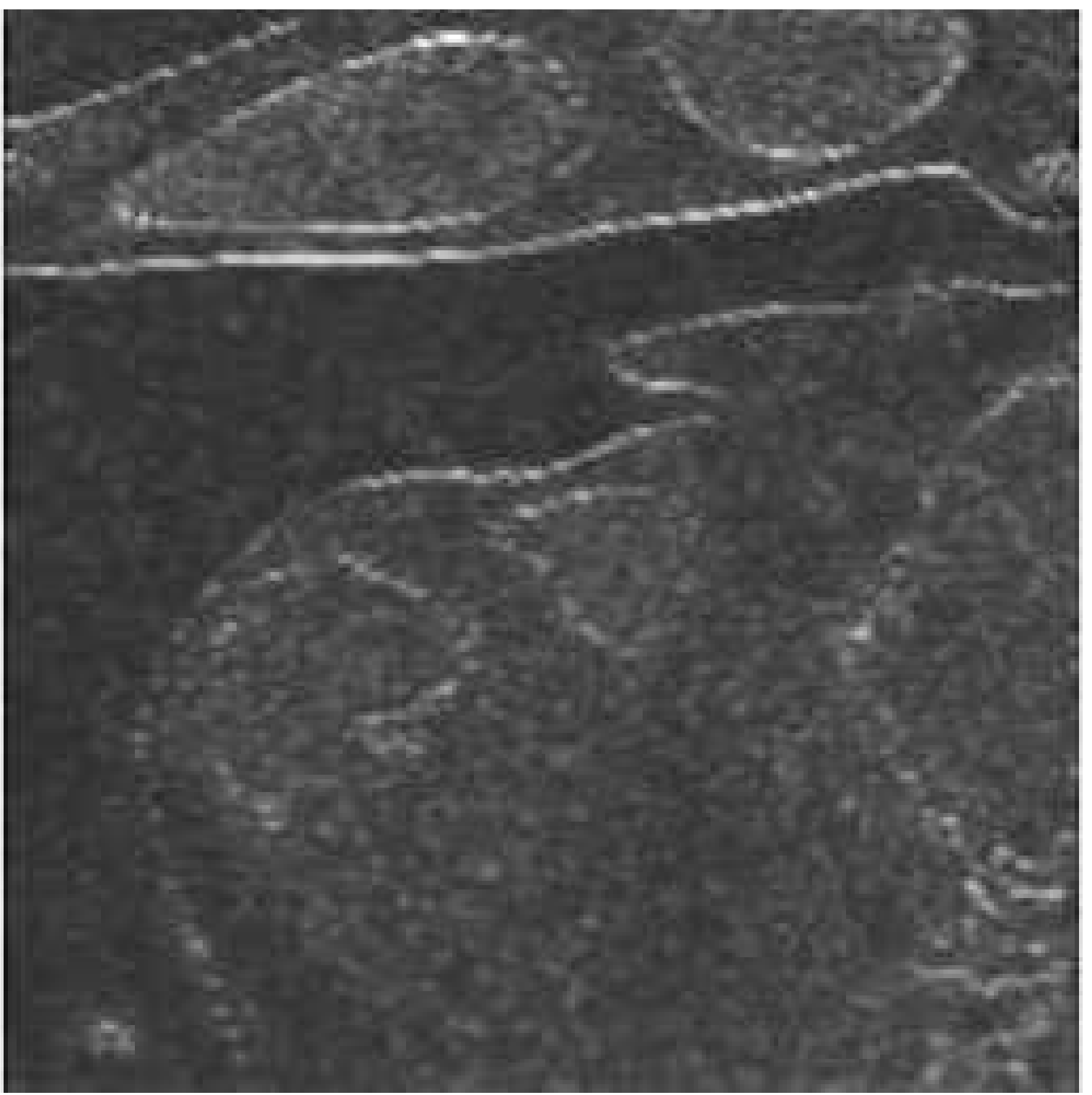} &
		\includegraphics[width = 0.24\textwidth]{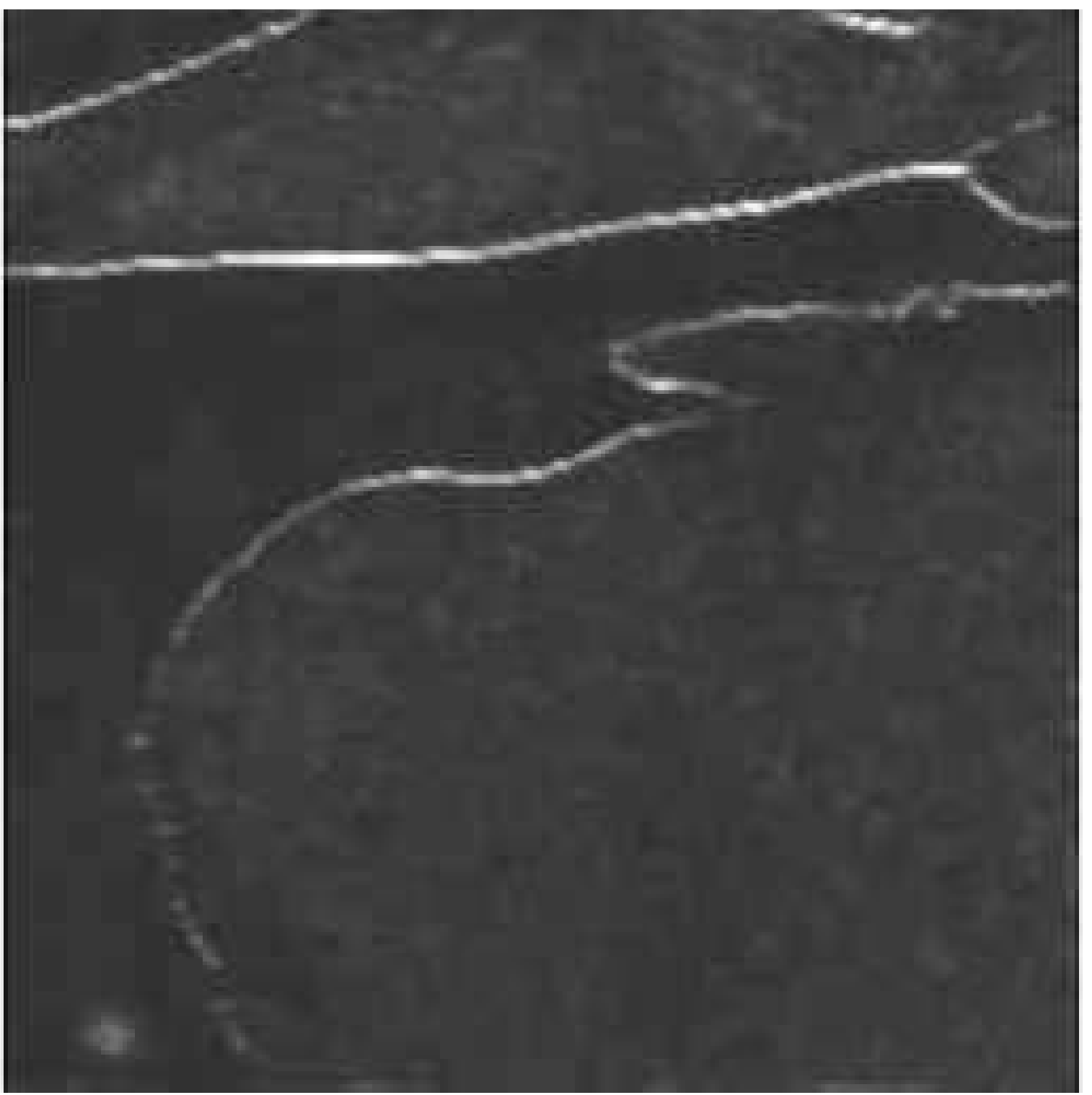} & \\    
		
		 & \small CNN-TV & \small CNN-NLM  & \small CNN-BM3D \\
          & \small $35.43$ dB &\small $32.6$ dB &\small $42.43$ dB &
		
	\end{tabular}   \\
	\smallskip
    \begin{tabular}{ c@{\hskip 0.001\textwidth}c@{\hskip 0.001\textwidth}c@{\hskip 0.001\textwidth}c@{\hskip 0.001\textwidth}c}
		\includegraphics[width = 0.24\textwidth]{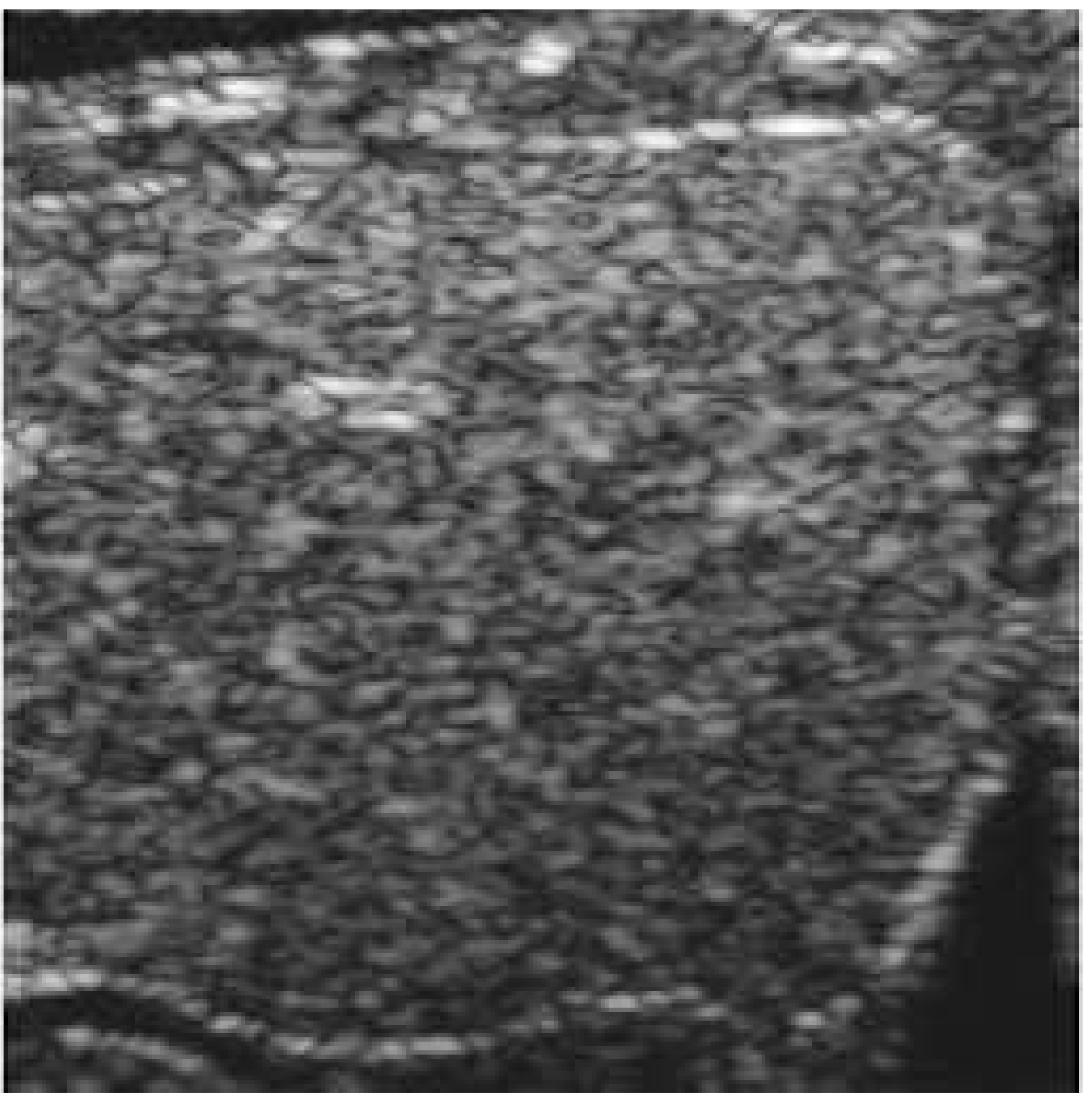} &
		\includegraphics[width = 0.24\textwidth]{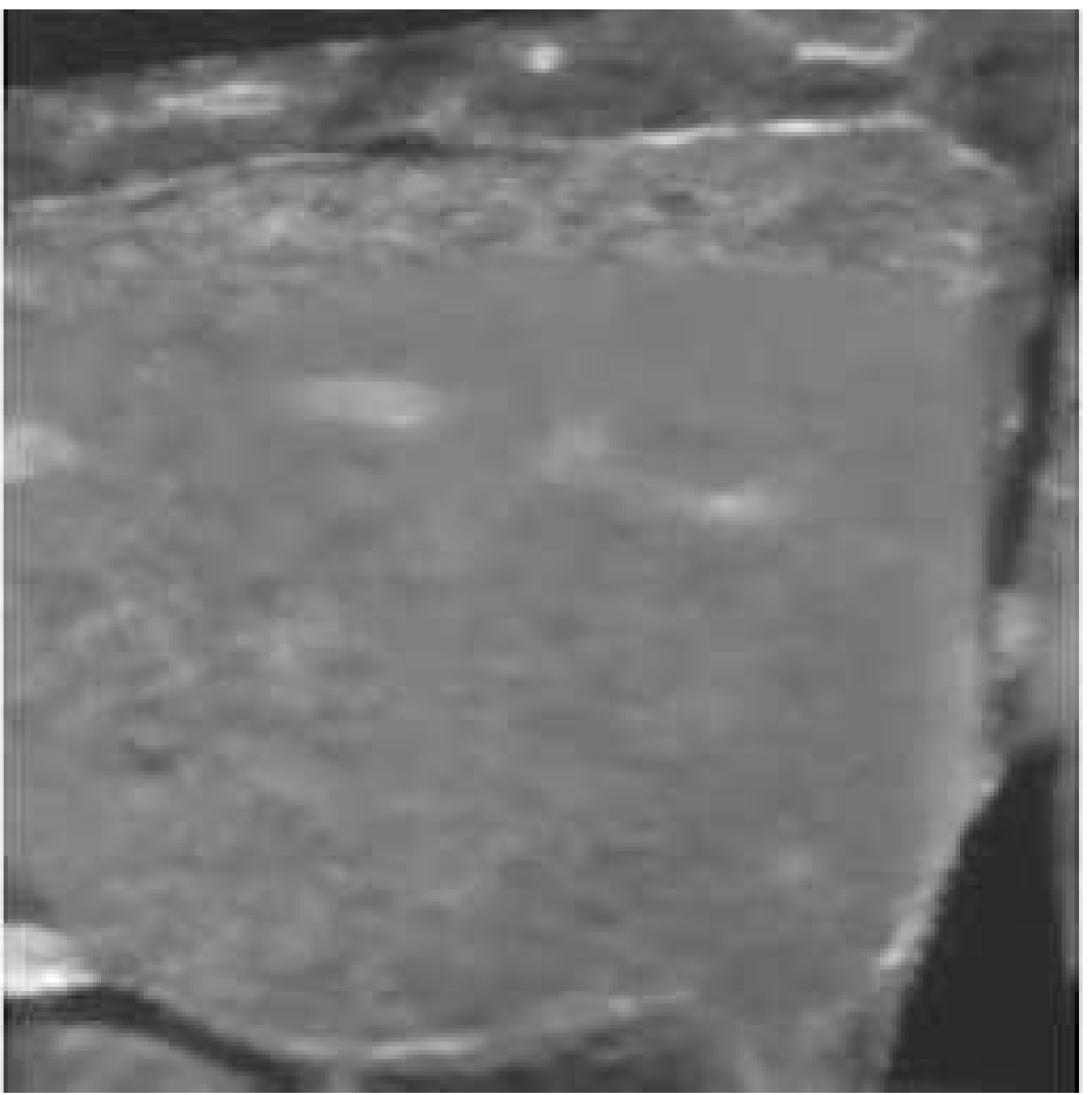} &
		\includegraphics[width = 0.24\textwidth]{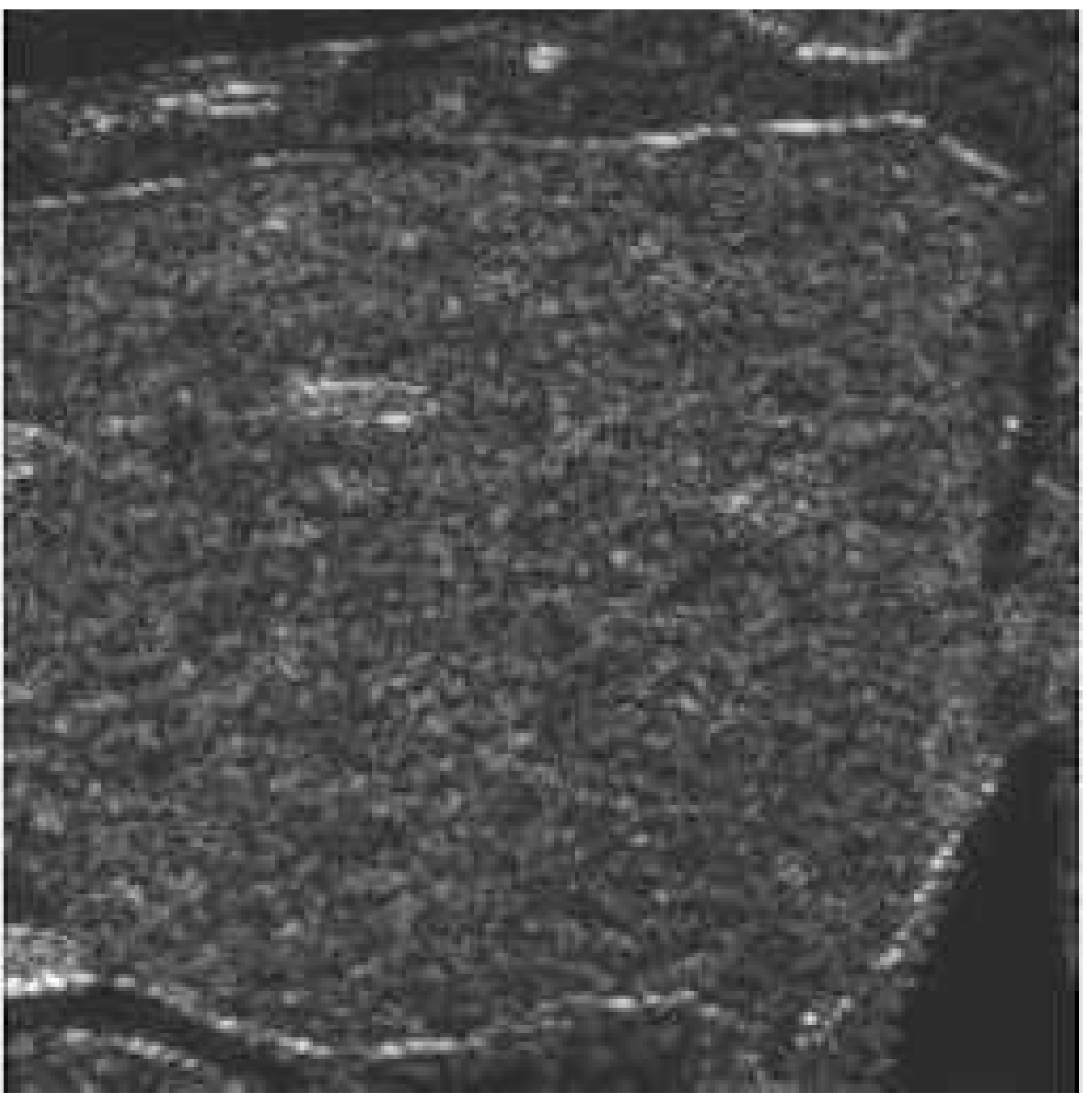} &
		\includegraphics[width = 0.24\textwidth]{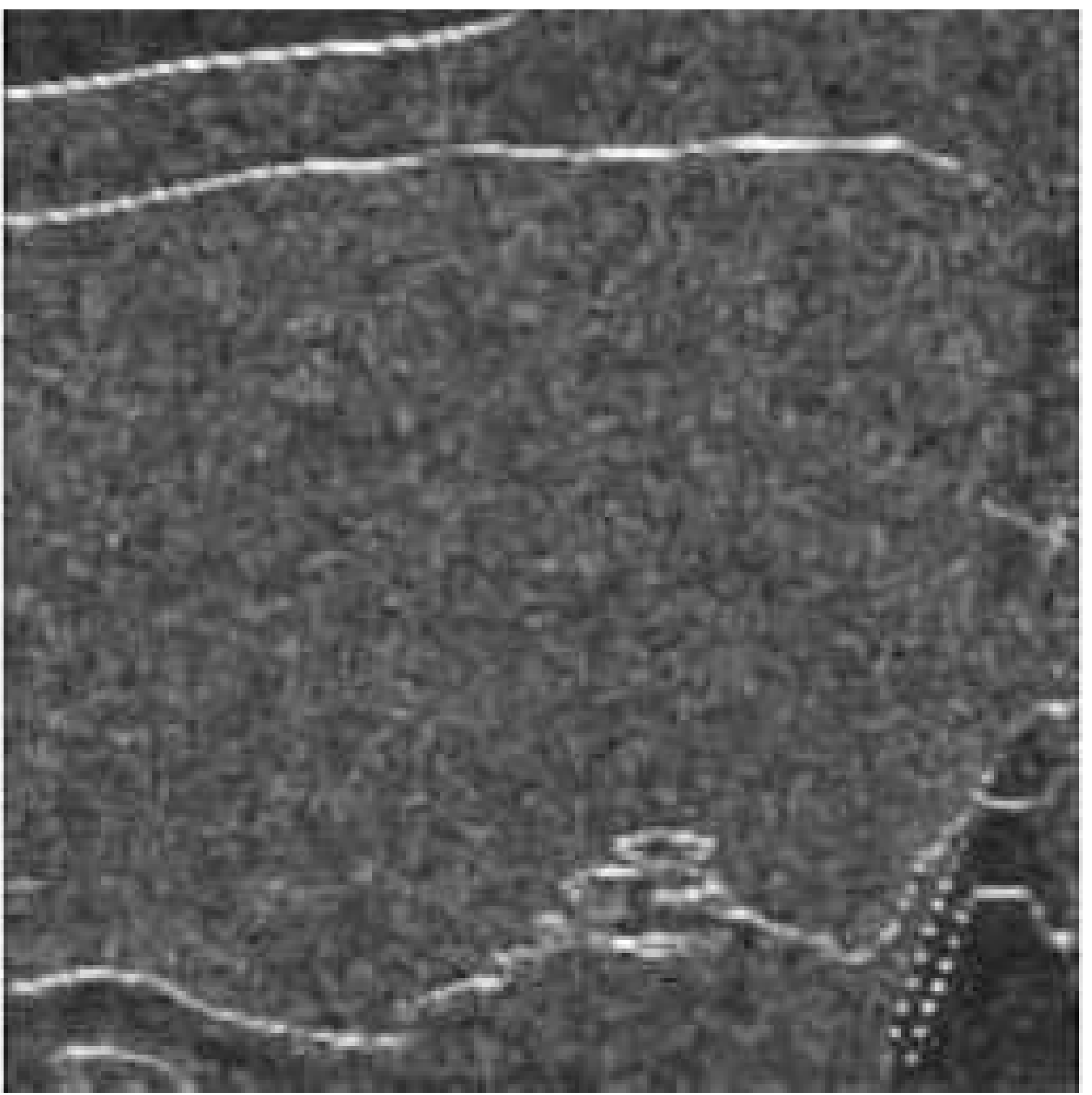} & \\  
		\small US image(b) & \small TV & \small NLM & \small BM3D \\
		
		 &
		\includegraphics[width = 0.24\textwidth]{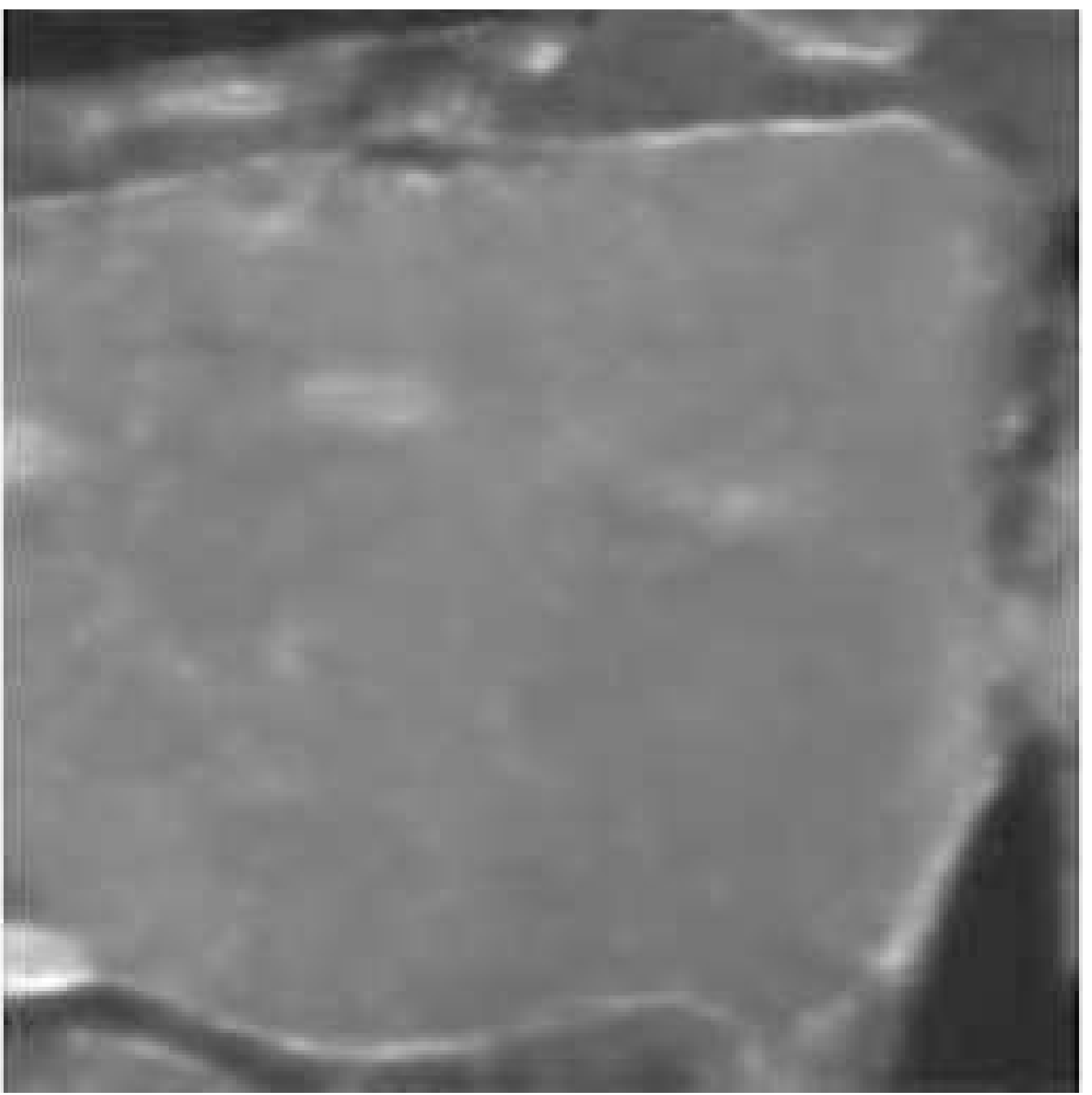} &
		\includegraphics[width = 0.24\textwidth]{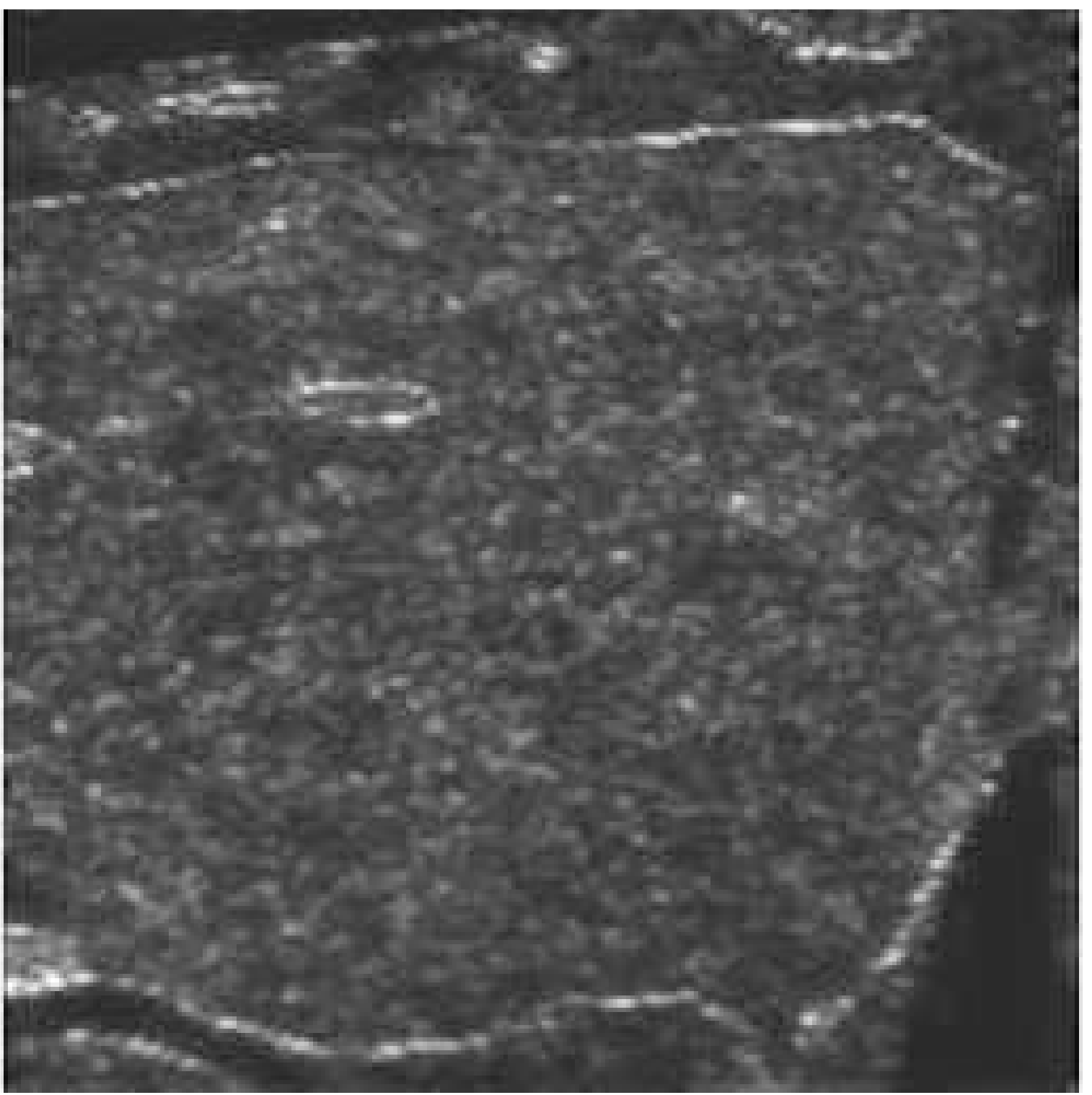} &
		\includegraphics[width = 0.24\textwidth]{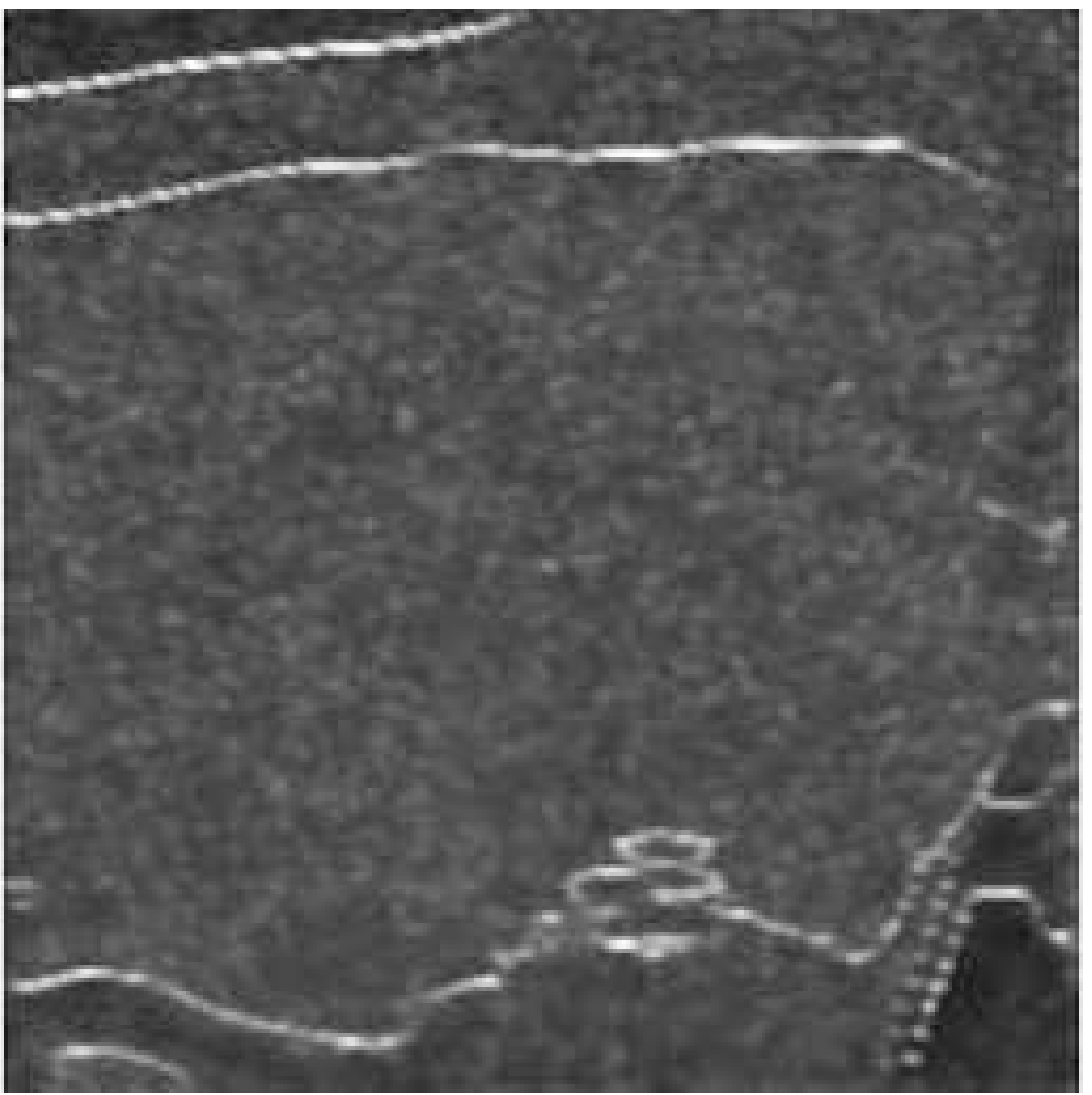} & \\    
		
		 & \small CNN-TV & \small CNN-NLM  & \small CNN-BM3D \\
          & \small  $31.08$ dB & \small $32.01$ dB & \small $32.26$ dB &
		
	\end{tabular} 
   \end{minipage}
   }
    \vspace{-0.2cm}
	\caption{\small \textbf{CNN approximation of different despeckling methods.}  Two examples of despeckling approximation are presented. Odd rows depict  conventional despeckling of the input US image in the leftmost column, while even rows display the CNN approximation. PSNR is reported for each CNN approximation with respect to the corresponding original despeckled image.}
	\label{Fig1}
\end{figure}

The run-time results for each algorithm are summarized in Table \ref{Table1} with their corresponding PSNR in Table \ref{Table2}. It can be observed that the CNN approximation of different despeckling algorithms gained a run-time speed-up in the range of $\times 4$ on the GPU and $\times 8-260$ on the CPU as compared to their off-the-shelf GPU/CPU counterparts, while maintaining an accurate approximation of the original despeckling output. A visual inspection of the results is displayed in Figure \ref{Fig1} showing that the CNN, upon training, can approximate any well-engineered time-consuming despeckling algorithms accurately while producing orders of magnitude of speed-up needed for real-time applications.

It is also important to note that the current results were obtained with minimal optimization of the network architecture which means that a more compact network might achieve similar results in a shorter time. \\

\begin{table}
\begin{center}
 \begin{tabular}{|c |c |c |c |c|} 
 \hline
  & BM3D & NLM & TV & CNN(ours) \\ [0.5ex] 
 \hline
CPU  & 390.7s & 400.3s & 12.8s & \textbf{1.5s}\\ 
\hline
GPU & - & - & 0.78s & \textbf{0.2s} \\  
 \hline
\end{tabular}
\caption{Run-time comparison of conventional and CNN-based to despeckling on a $256\times256\times64$ US volume}
\label{Table1}
\end{center}
\end{table}

\begin{table}
\begin{center}
 \begin{tabular}{|c |c |c |c |c|} 
 \hline
       & BM3D & NLM & TV \\ [0.5ex] 
 \hline
 PSNR  & 34.40dB & 35.33dB & 41.66dB\\ 
 \hline
\end{tabular}
\vspace*{-2mm}
\caption{Average PSNR of despeckling approximation by CNN calculated with respect to the conventional despeckling output image. }
\label{Table2}
\end{center}
\vspace{-0.5cm}
\end{table}


\vspace{-4mm}
\subsection{Reconstructing `CT-quality' images} 
\label{ssec:IQ2CT}
\textbf{Datasets.} The CT image patches were extracted from the TCIA datasets\cite{TCIA} with corresponding ultrasound IQ images generated using the simulator described in Section \ref{ssec:data}.\\
\\
\textbf{Training.} Using the network architecture proposed in Section \ref{ssec:NetworkArchitecture}, training was performed on a dataset comprising of 13,860 patches from ultrasound IQ images ($2$ channels) as input and the corresponding CT patches as the output. Training was run for $160K$ mini-batches for $4$ hours on an NVIDIA Titan X GPU. \\
\\
\textbf{Results.} Results are presented in Figure 2. We compare the performance of the reconstructed CT images to conventional and CNN-approximated despeckling methods, and for all the images, the PSNR is presented with respect to the ground truth CT(columns $3$ and $4$ in Figure 2). We can observe both visually and quantitatively that our reconstructed CT images look closer to the ground truth CT when compared to the despeckled ones.

Additionally, in order to verify that our network does not overfit the dataset, and generalizes well to previously unseen data, we tested it on a different CT dataset obtained from TCIA. The average PSNR of the reconstructed CT-quality images with respect to ground truth CT was $23.79$ dB, which resembles the performance on the training set. 


We can observe in our results (column $1$ and $5$ in Figure $2$) that the trained network is able to recover the overall anatomy and the contrast accurately, except for the CT-specific noise, a feature that CT and ultrasound imaging do not share. A possible explanation could be that the network learns the common manifold of the CT and US domains, but not intrinsic features like modality-specific noise.     

\begin{figure}[t]
\makebox[0pt][l]{%
 \begin{minipage}[]{\linewidth}
	\begin{tabular}{c@{\hskip 0.001\textwidth}c@{\hskip 0.001\textwidth}c@{\hskip 0.001\textwidth}c@{\hskip 0.001\textwidth}c}    
    
		CT GT & US Image  & TV & CNN-TV & CNN-CT \\
		\vspace*{-2mm}
		\includegraphics[width = 0.199\textwidth]{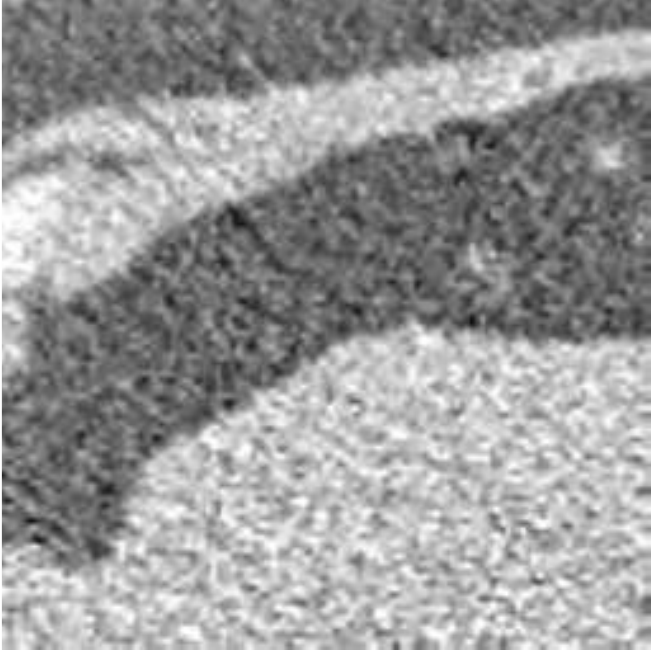} &
        \includegraphics[width = 0.199\textwidth]{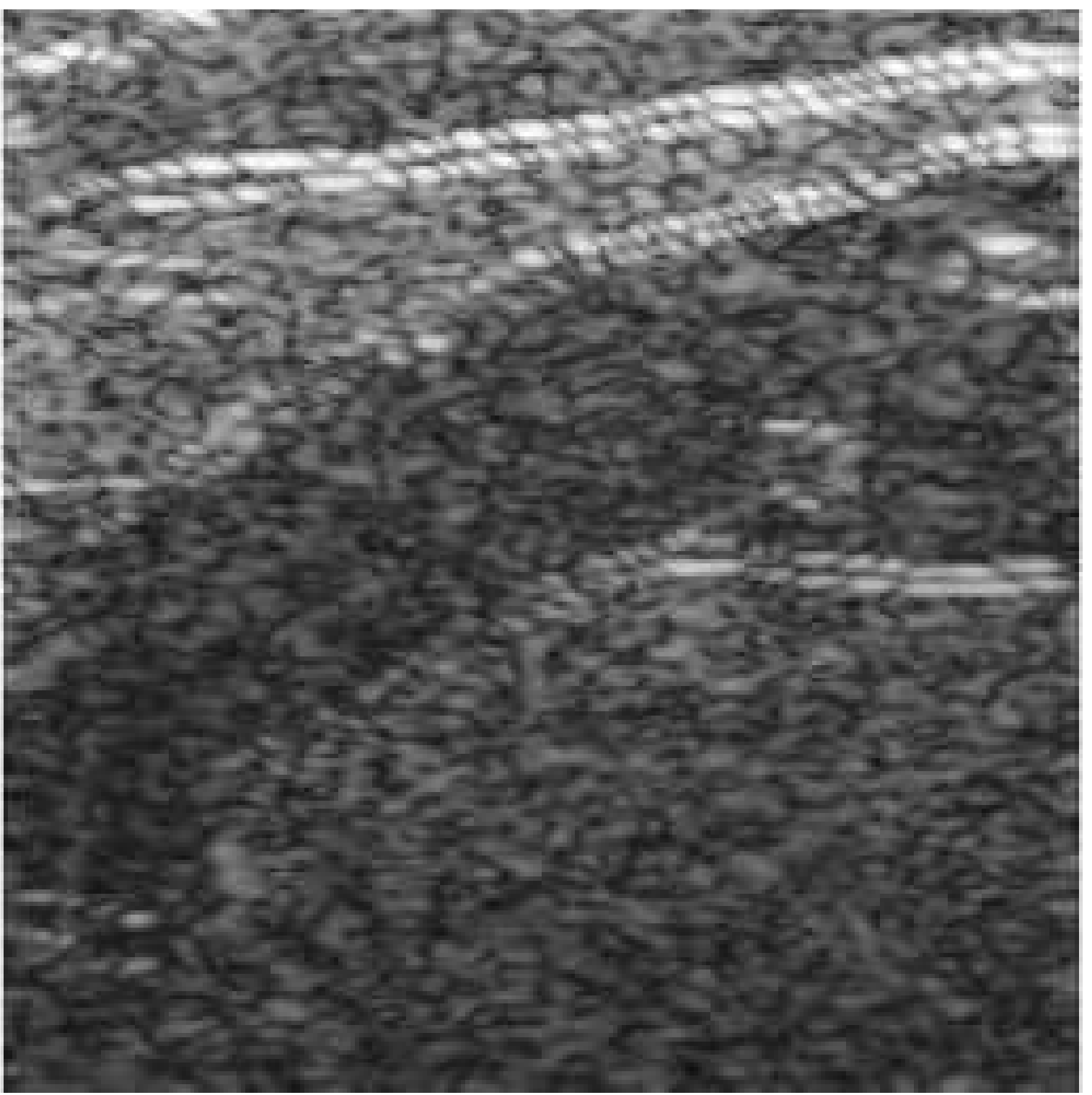} &
		\includegraphics[width = 0.199\textwidth]{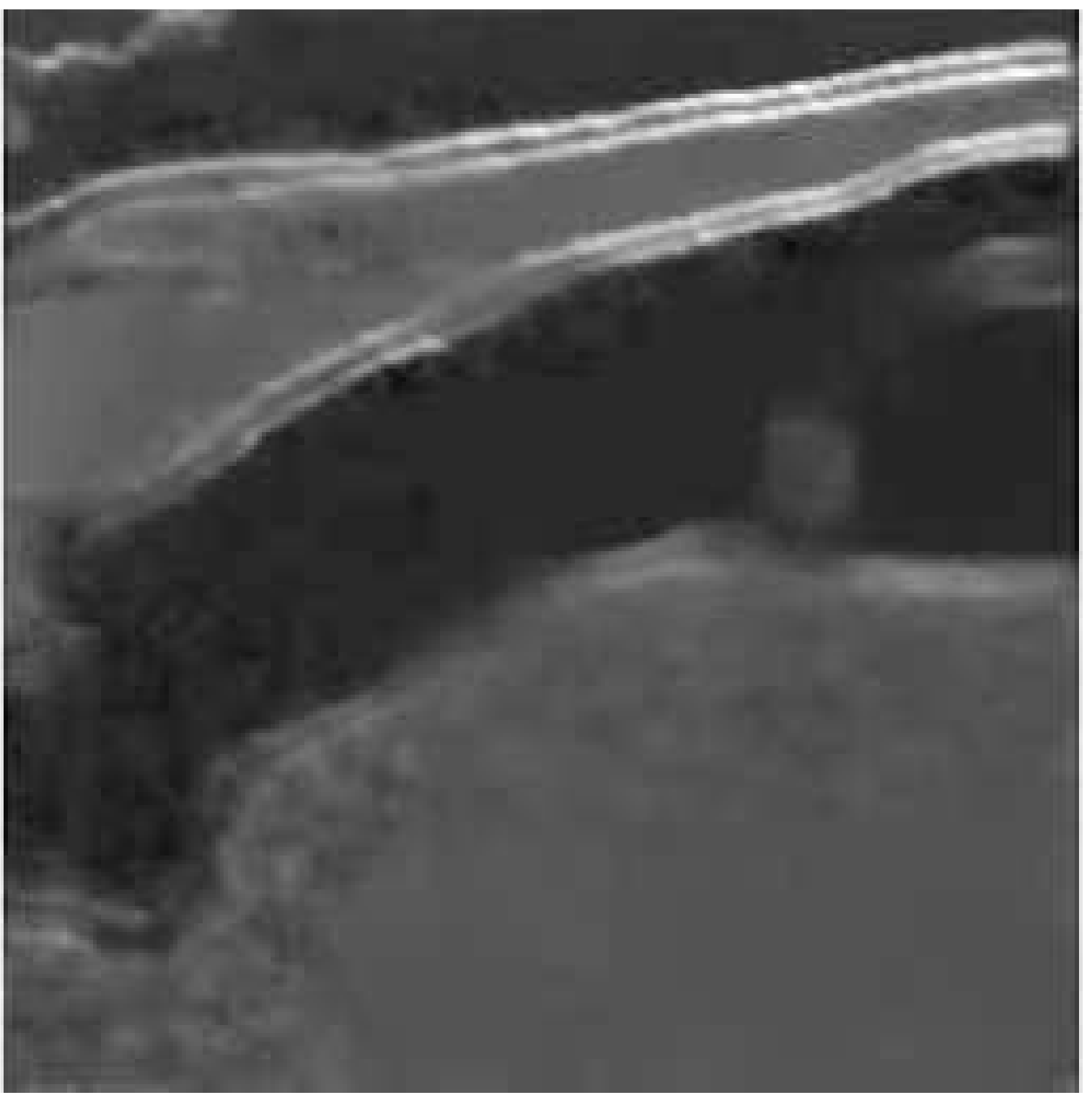} &
		\includegraphics[width = 0.199\textwidth]{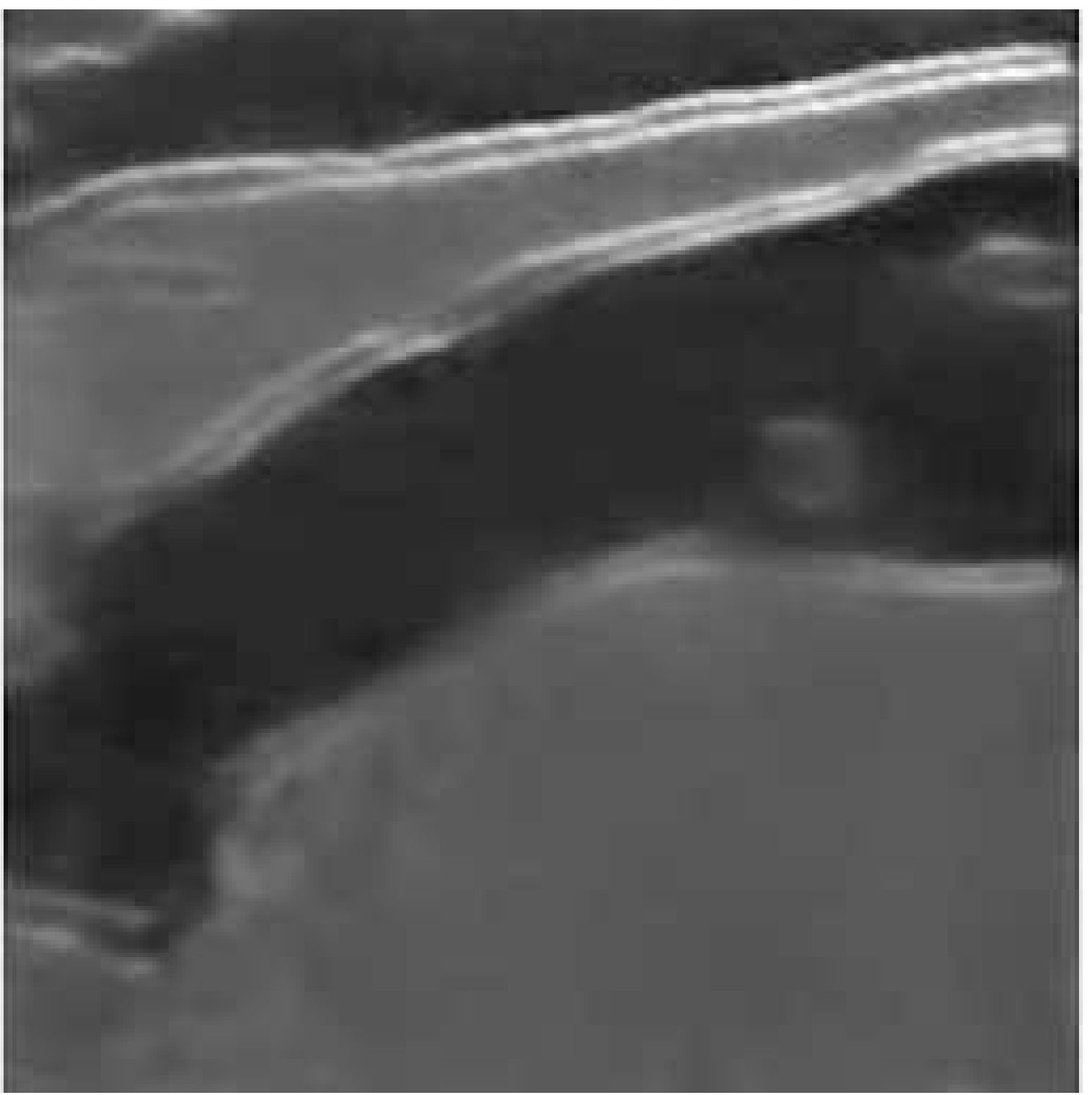} &
		\includegraphics[width = 0.199\textwidth]{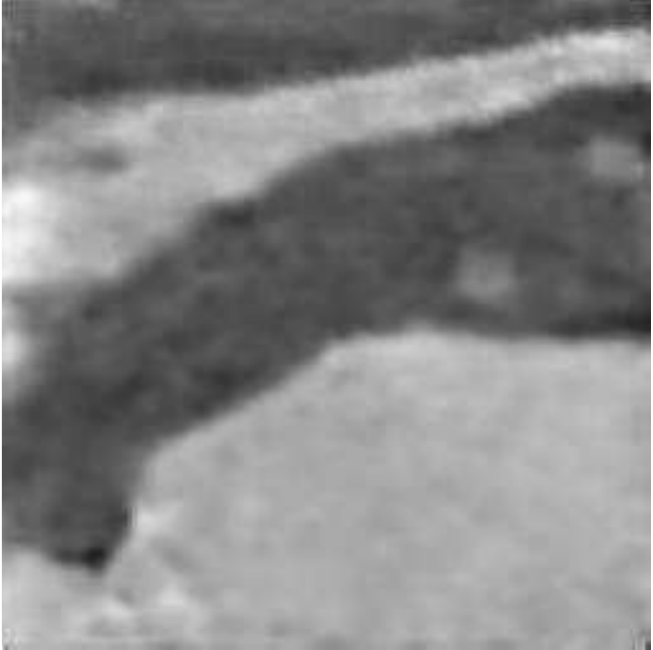} \\ 
		& & \small $5.32$ dB & \small $5.28$ dB & \small $18.22$ dB 
       \vspace*{-1mm}\\  
        \vspace*{-2mm} 
        \includegraphics[width = 0.199\textwidth]{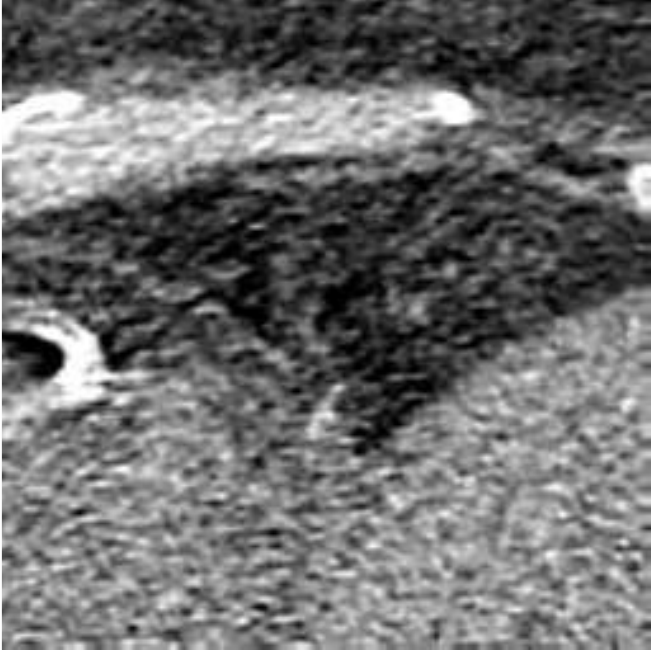} &
		\includegraphics[width = 0.199\textwidth]{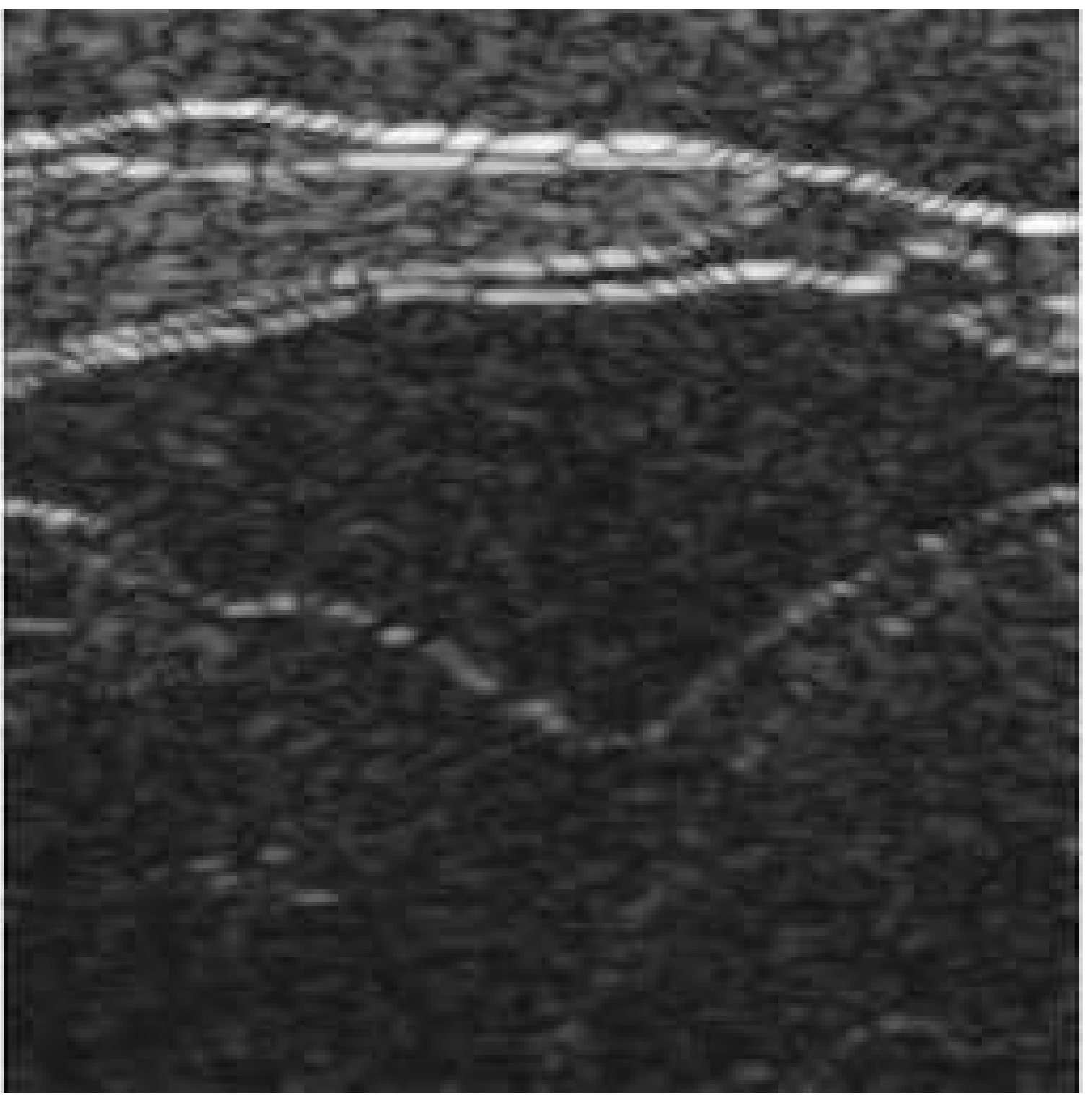} &
		\includegraphics[width = 0.199\textwidth]{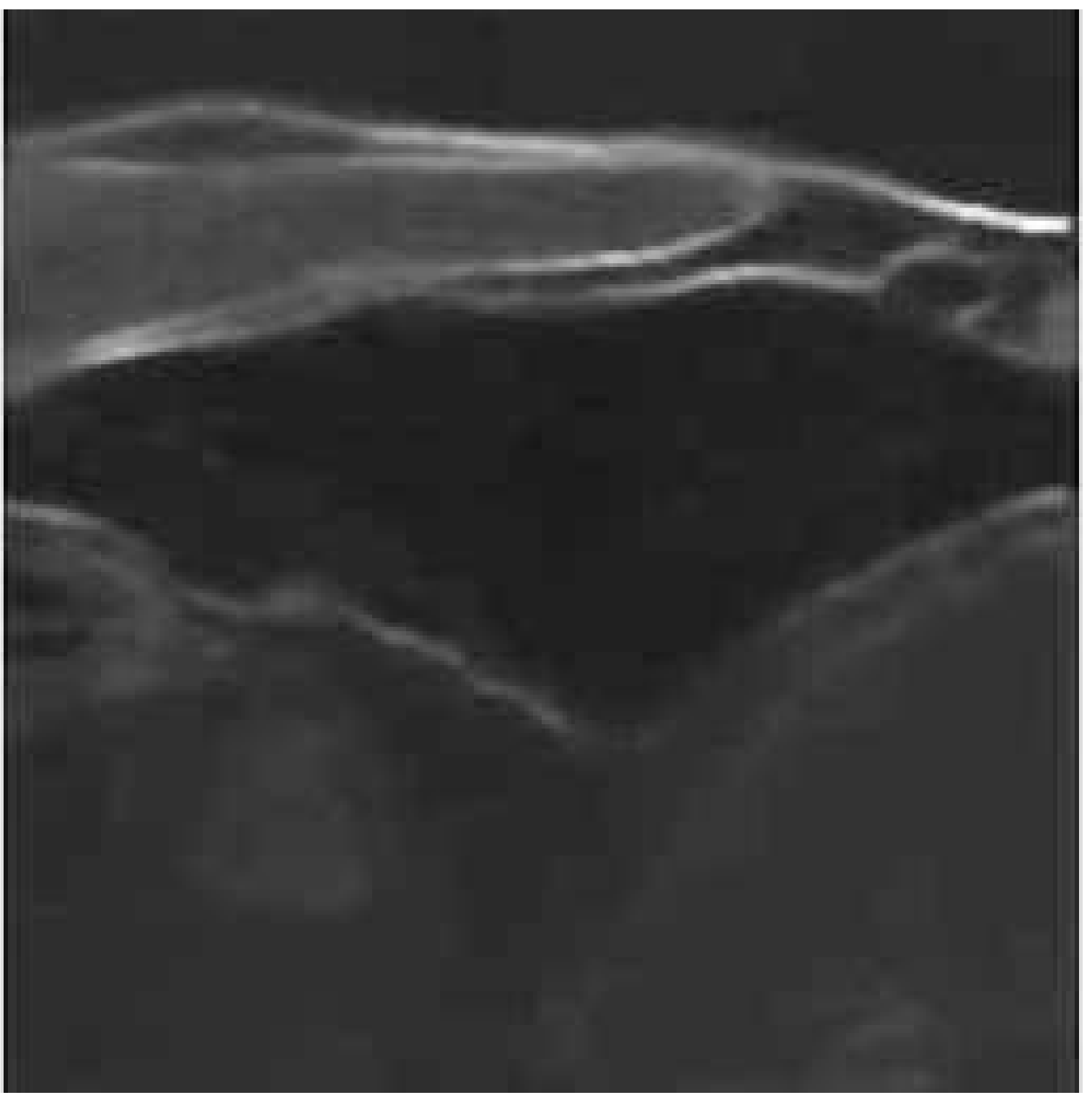} &
		\includegraphics[width = 0.199\textwidth]{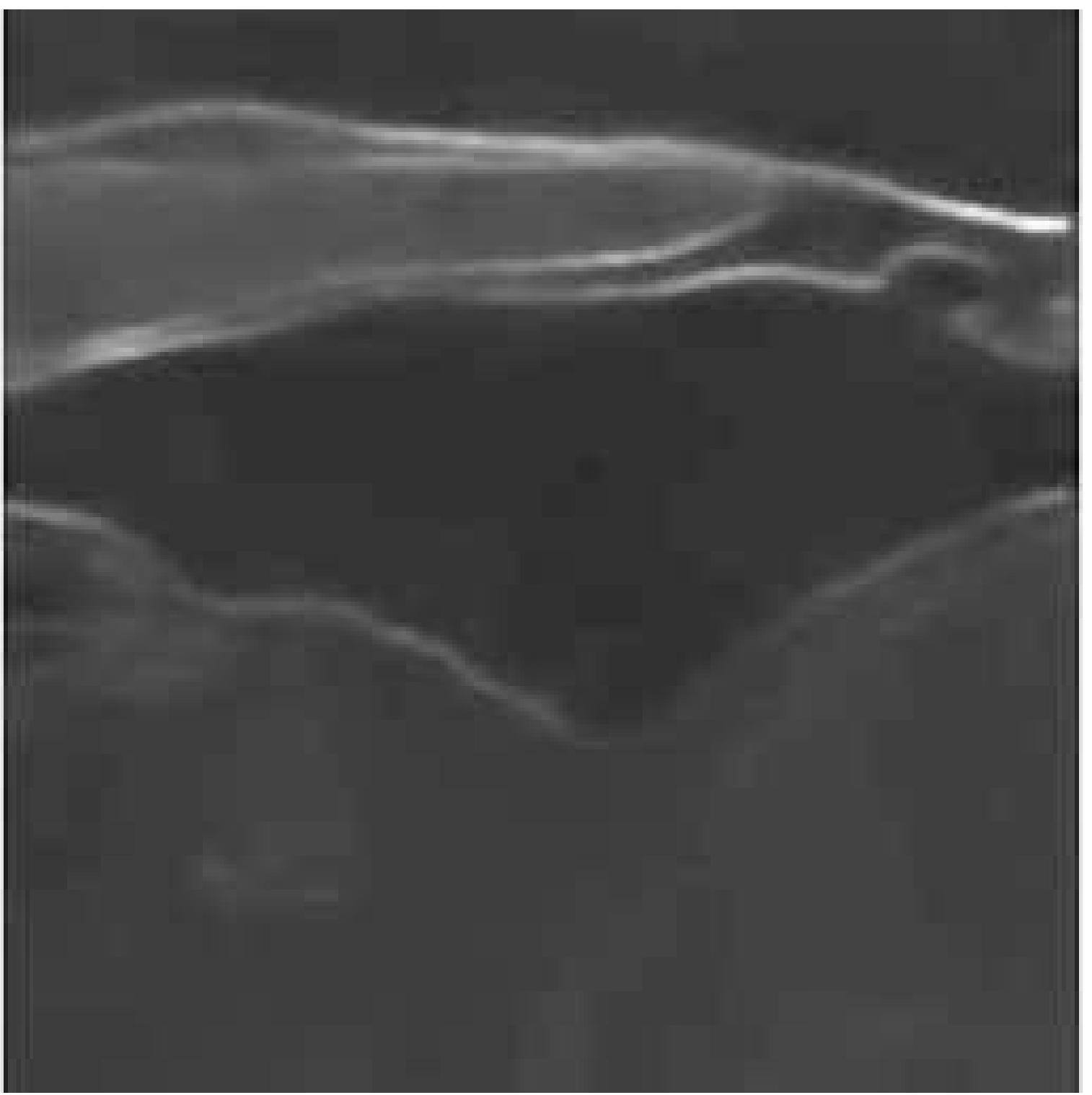} &
		\includegraphics[width = 0.199\textwidth]{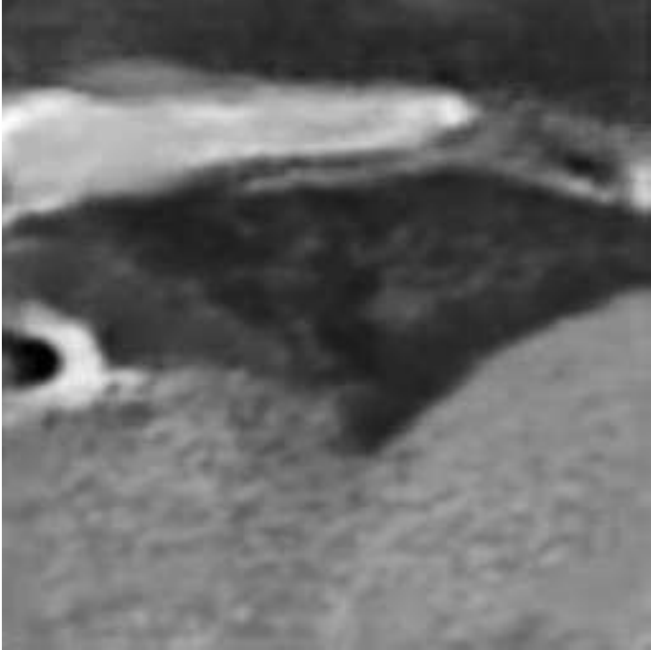} \\
		& & \small $6.99$ dB & \small $6.93$ dB & \small $18.5$ dB \vspace*{-1mm}\\
		\vspace*{-2mm}
        \includegraphics[width = 0.199\textwidth]{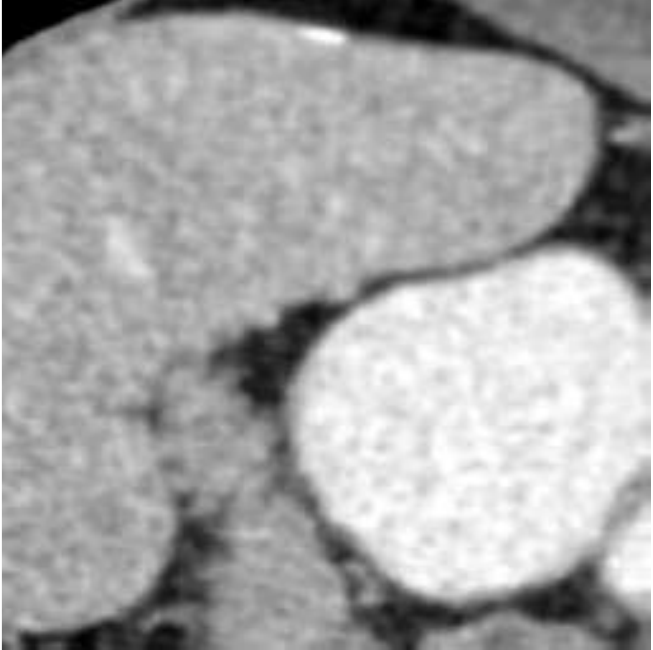} &
		\includegraphics[width = 0.199\textwidth]{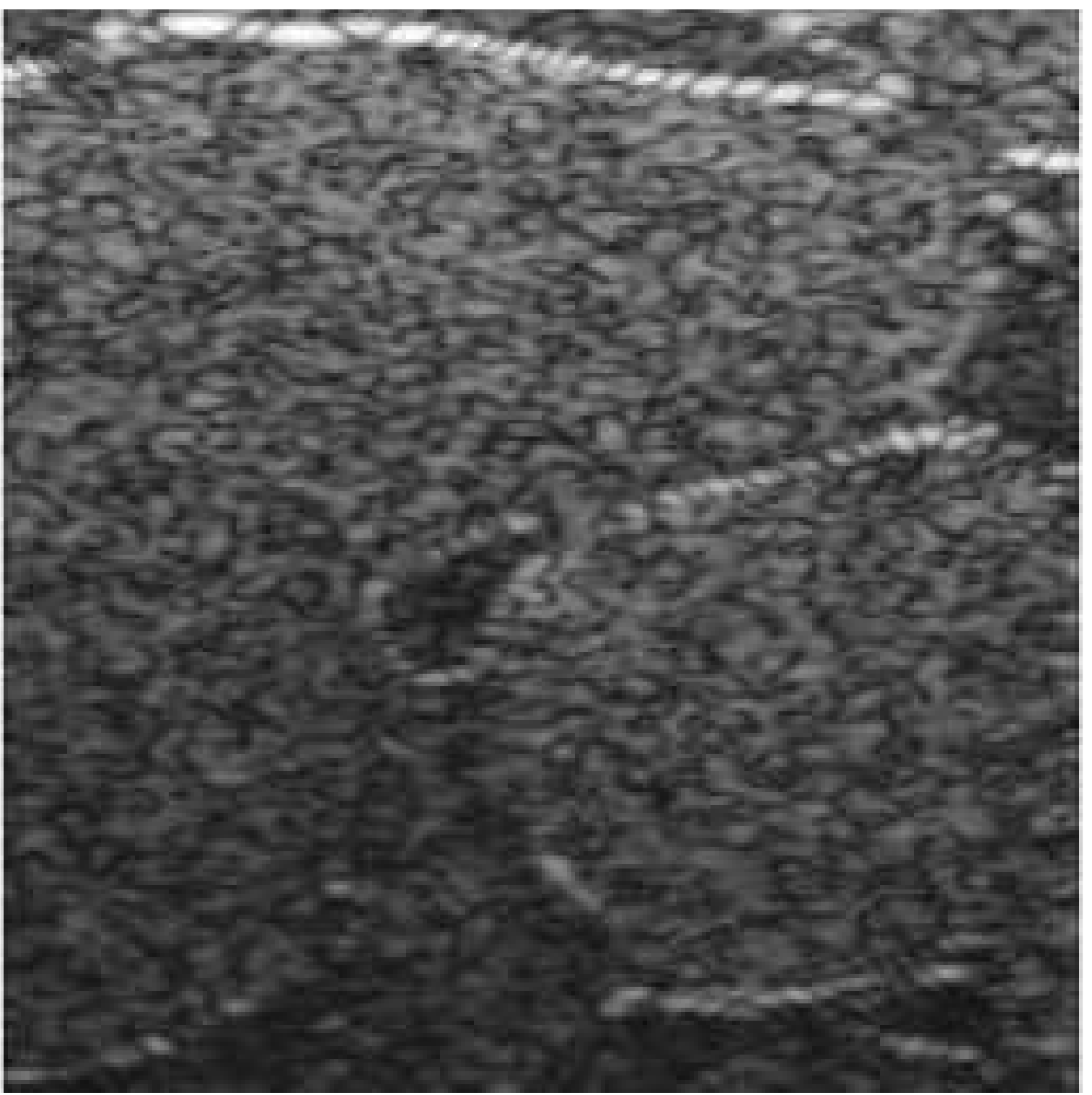} &
		\includegraphics[width = 0.199\textwidth]{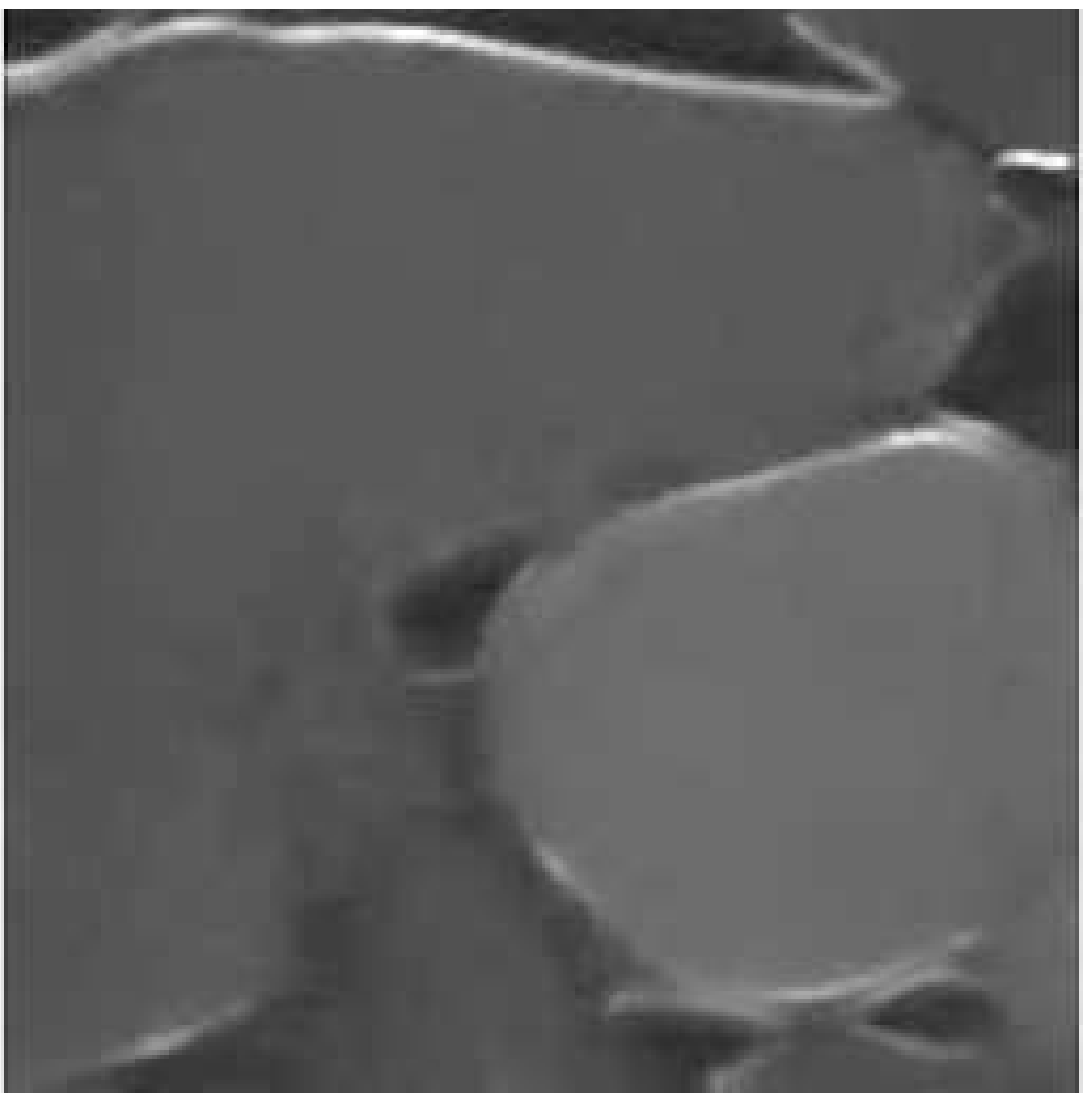} &
		\includegraphics[width = 0.199\textwidth]{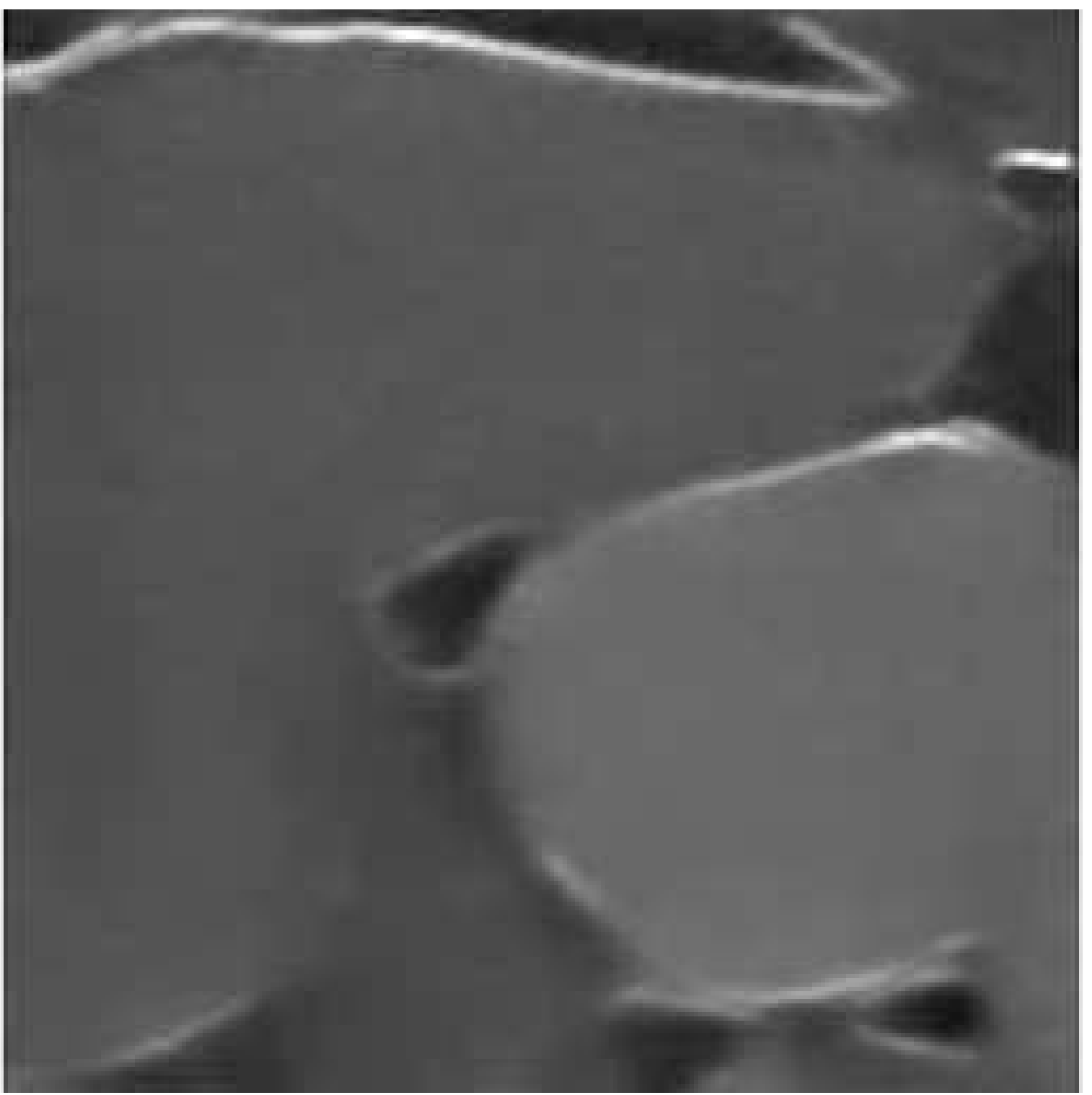} &
		\includegraphics[width = 0.199\textwidth]{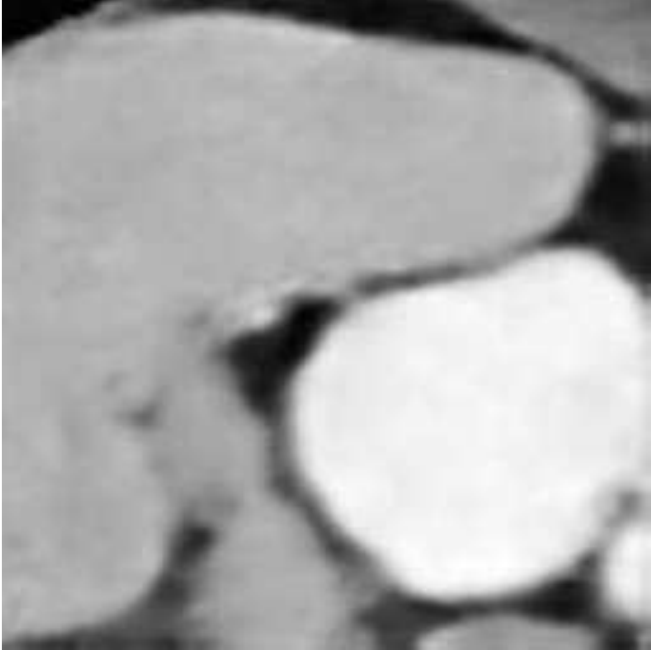}  \\ 
	    & &  \small $4.42$ dB & \small $4.68$ dB & \small $24.57$ dB \vspace*{-1mm}\\ 
		\vspace*{-2mm}		
        \includegraphics[width = 0.199\textwidth]{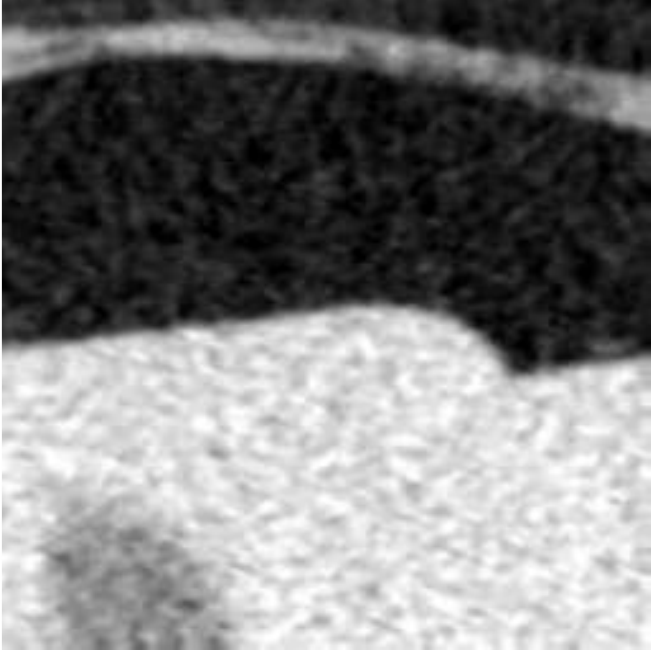} &
		\includegraphics[width = 0.199\textwidth]{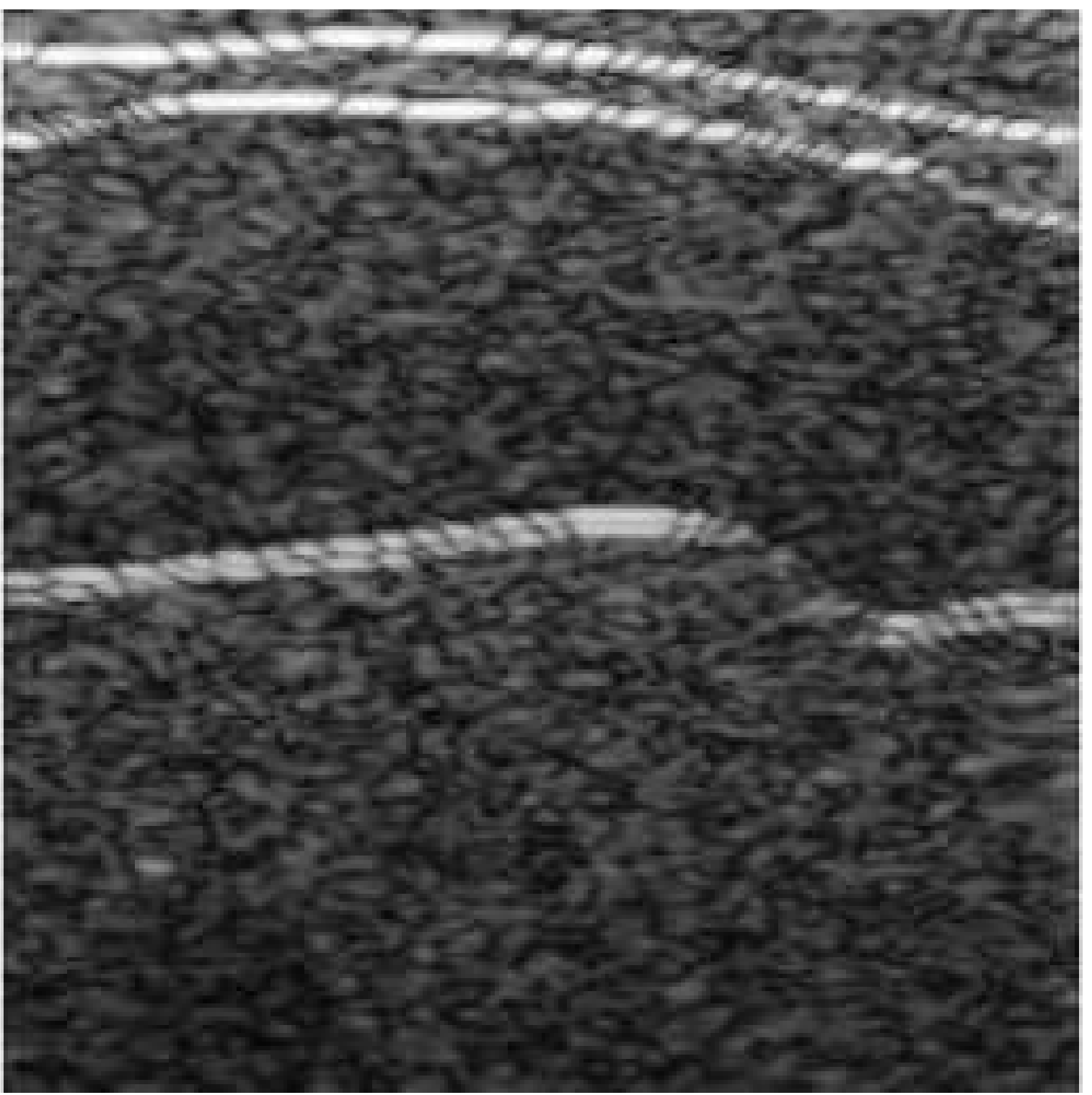} &
		\includegraphics[width = 0.199\textwidth]{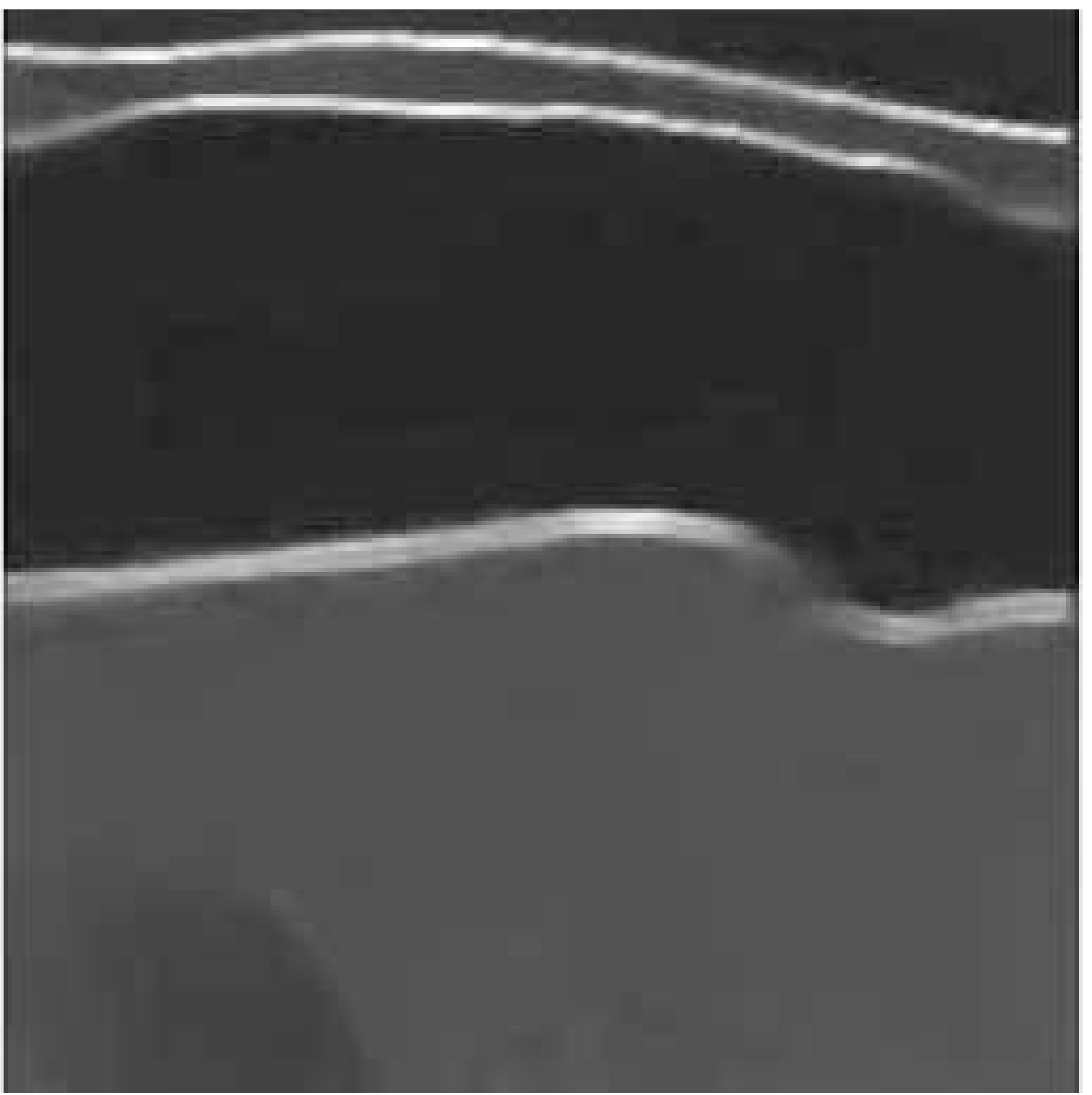} &
		\includegraphics[width = 0.199\textwidth]{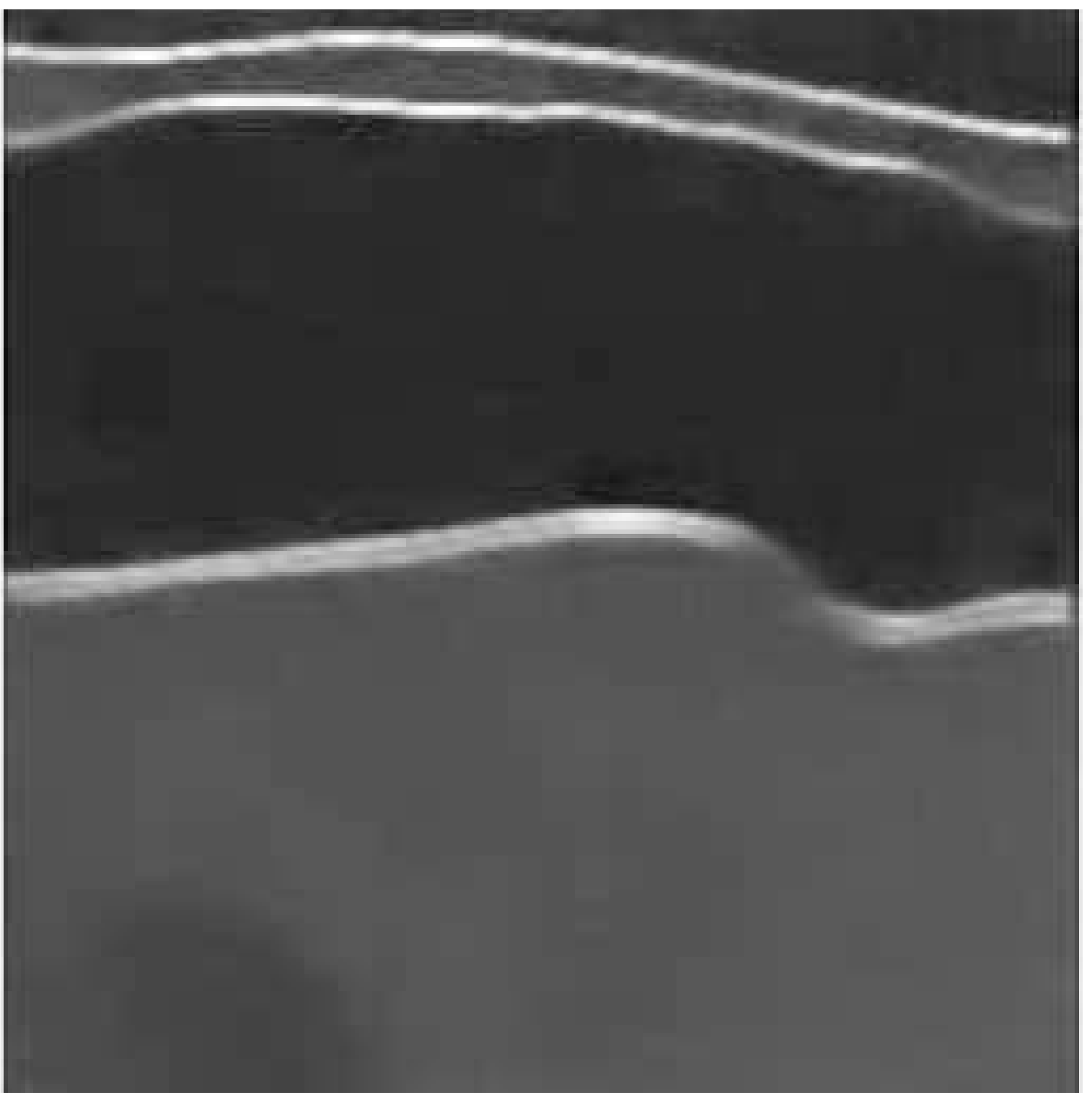} &
		\includegraphics[width = 0.199\textwidth]{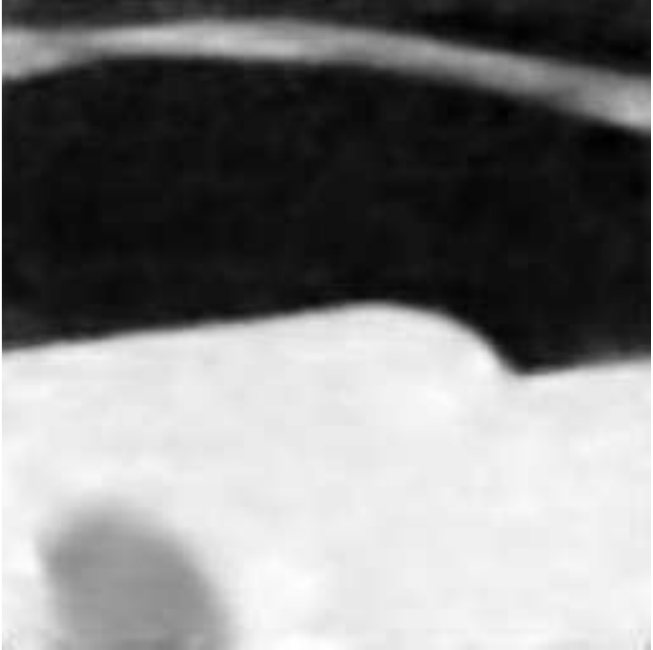}  \\ 
		& & \small $5.17$ dB & \small $5.38$ dB & \small $23.37$ dB \vspace*{-1mm}\\		
        \vspace*{-2mm}
        \includegraphics[width = 0.199\textwidth]{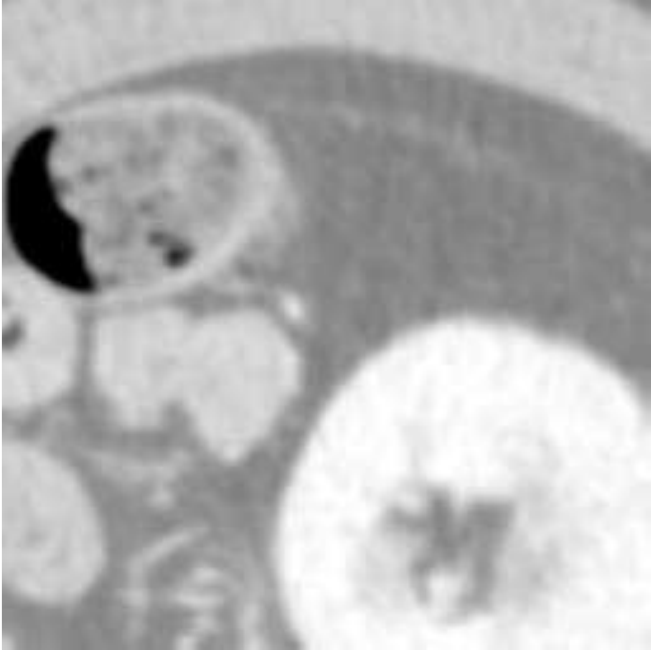} &
		\includegraphics[width = 0.199\textwidth]{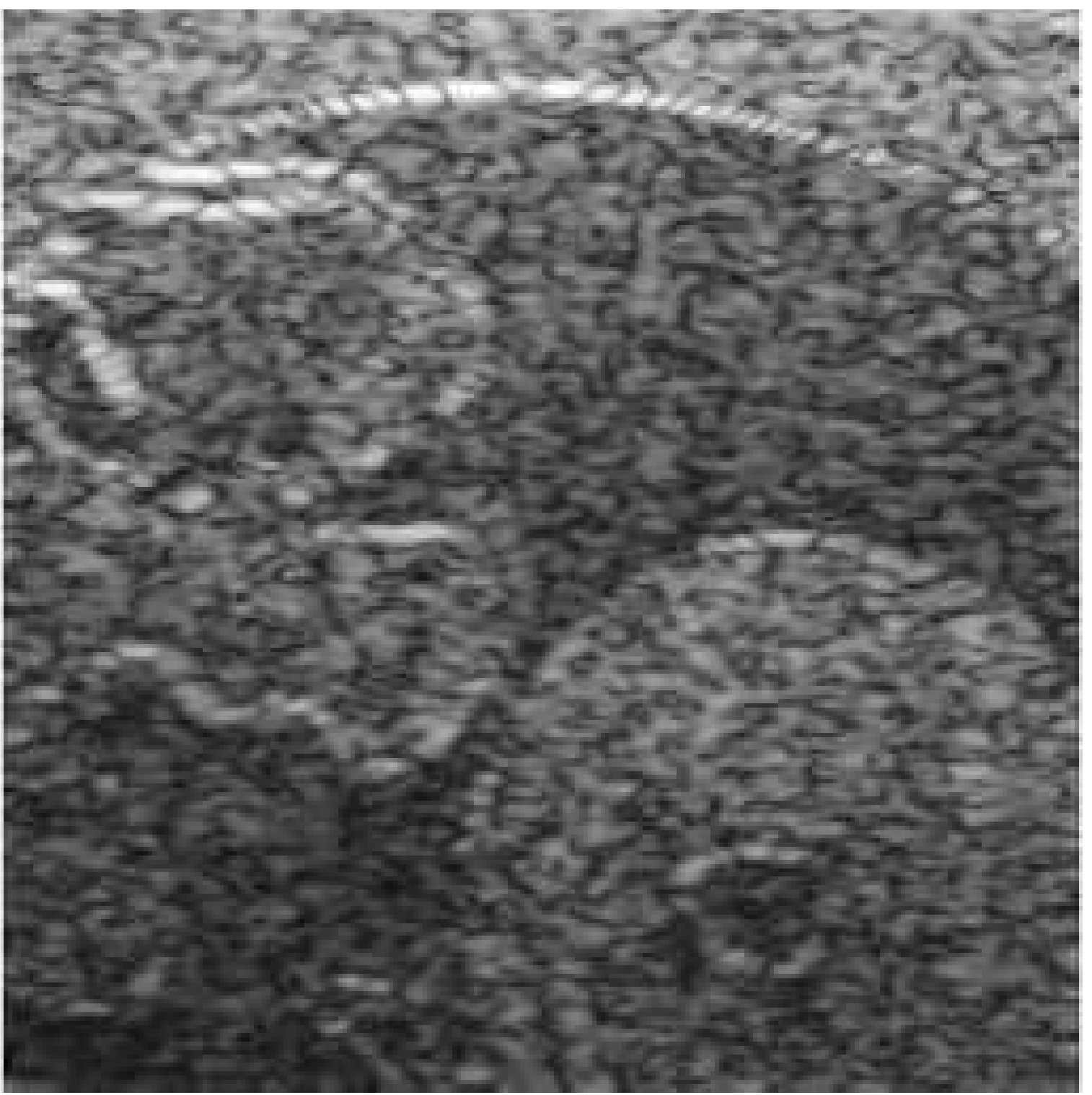} &
		\includegraphics[width = 0.199\textwidth]{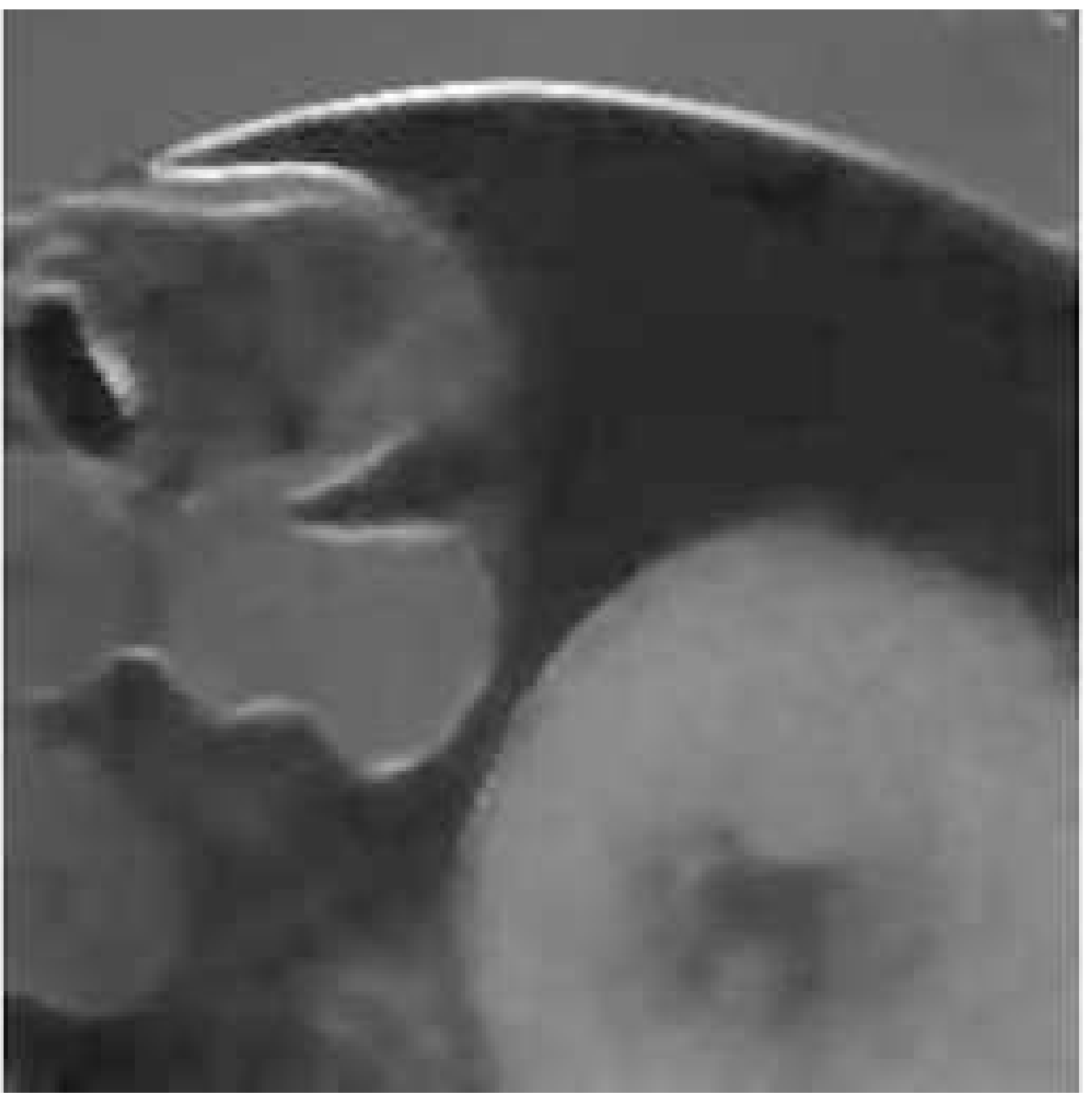} &
		\includegraphics[width = 0.199\textwidth]{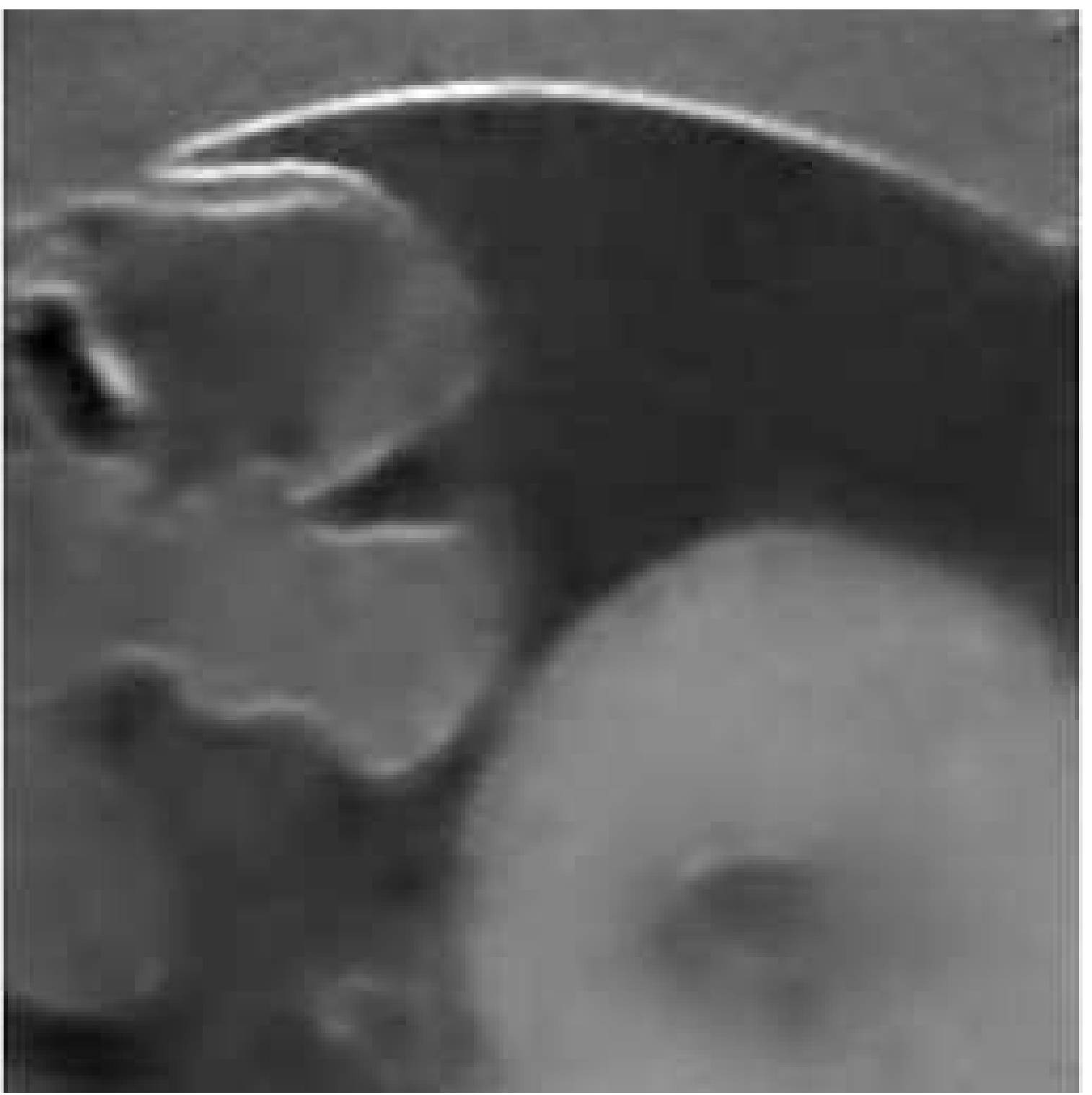} &
		\includegraphics[width = 0.199\textwidth]{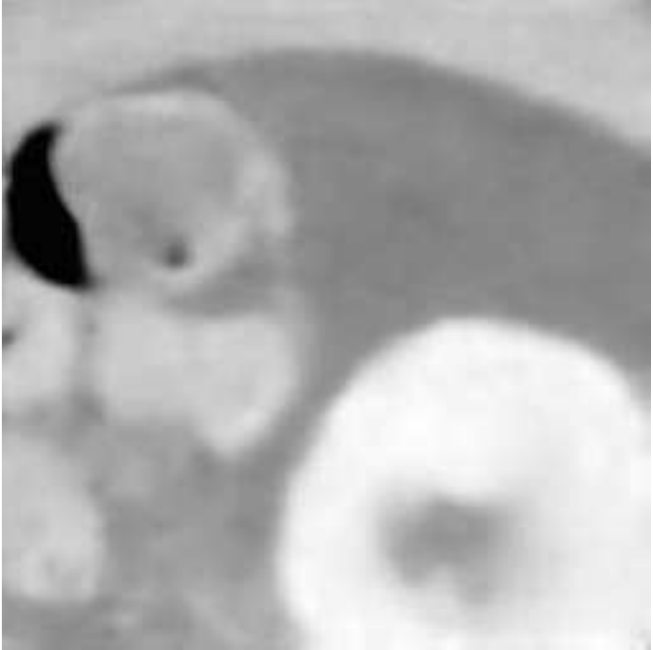}  \\ 
		&  & \small $4.21$ dB & \small $4.57$ dB & \small $27.64$ dB \vspace*{-1mm}\\
		\vspace*{-2mm}		
        \includegraphics[width = 0.199\textwidth]{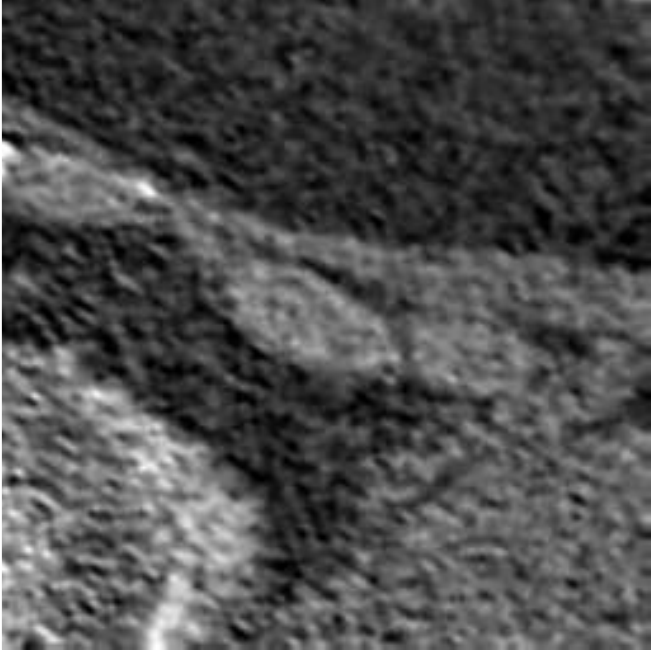} &
		\includegraphics[width = 0.199\textwidth]{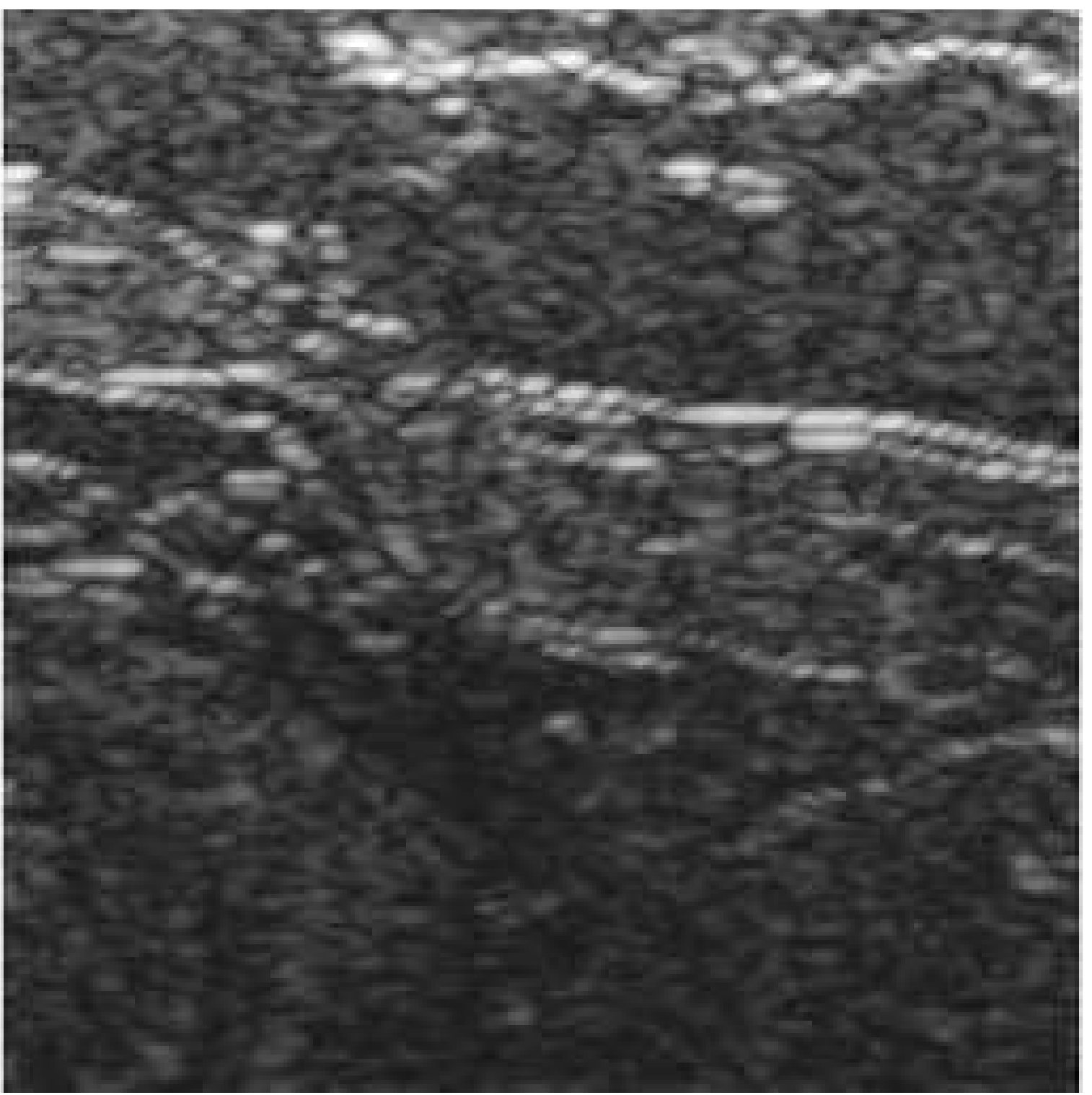} &
		\includegraphics[width = 0.199\textwidth]{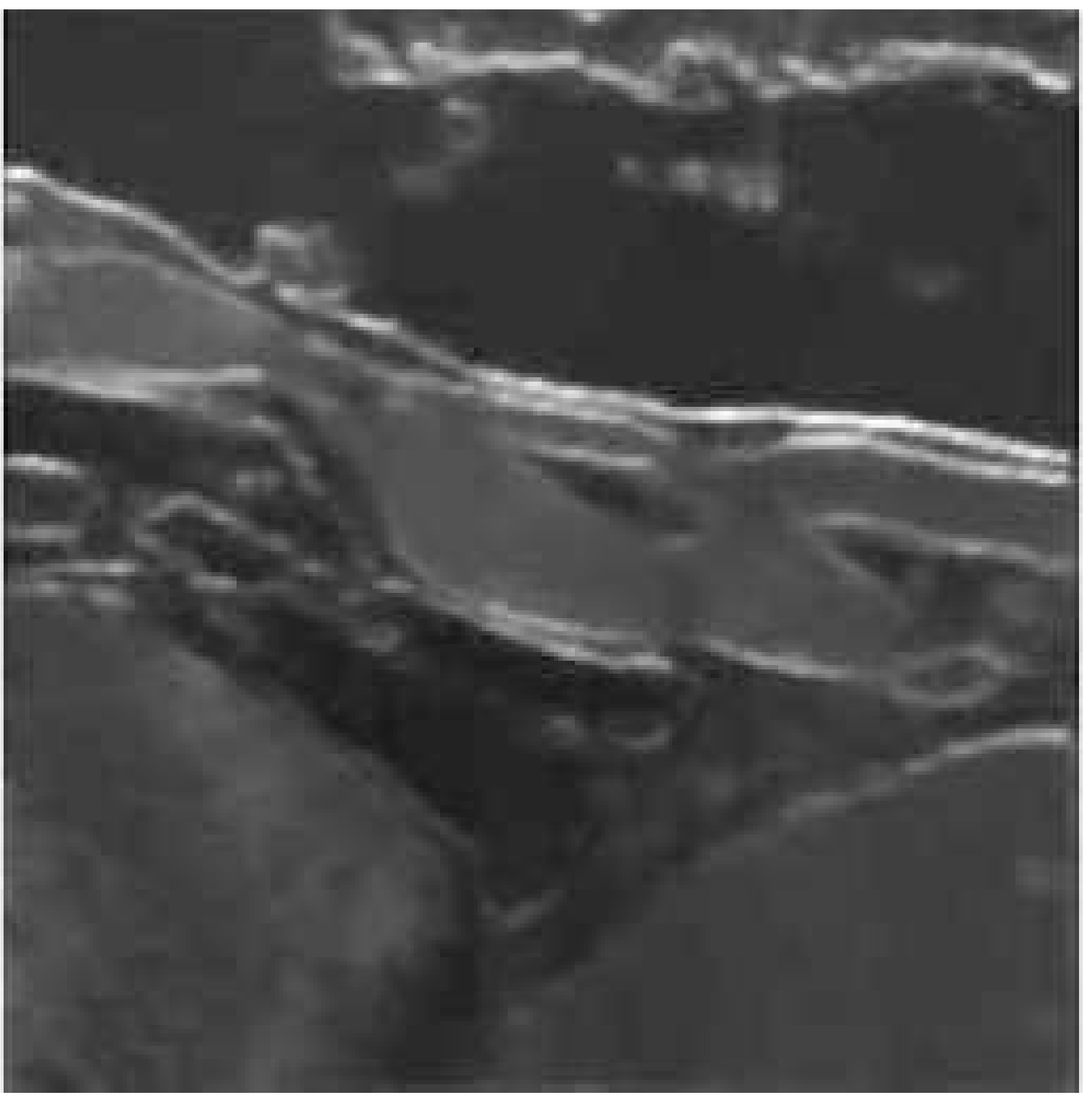} &
		\includegraphics[width = 0.199\textwidth]{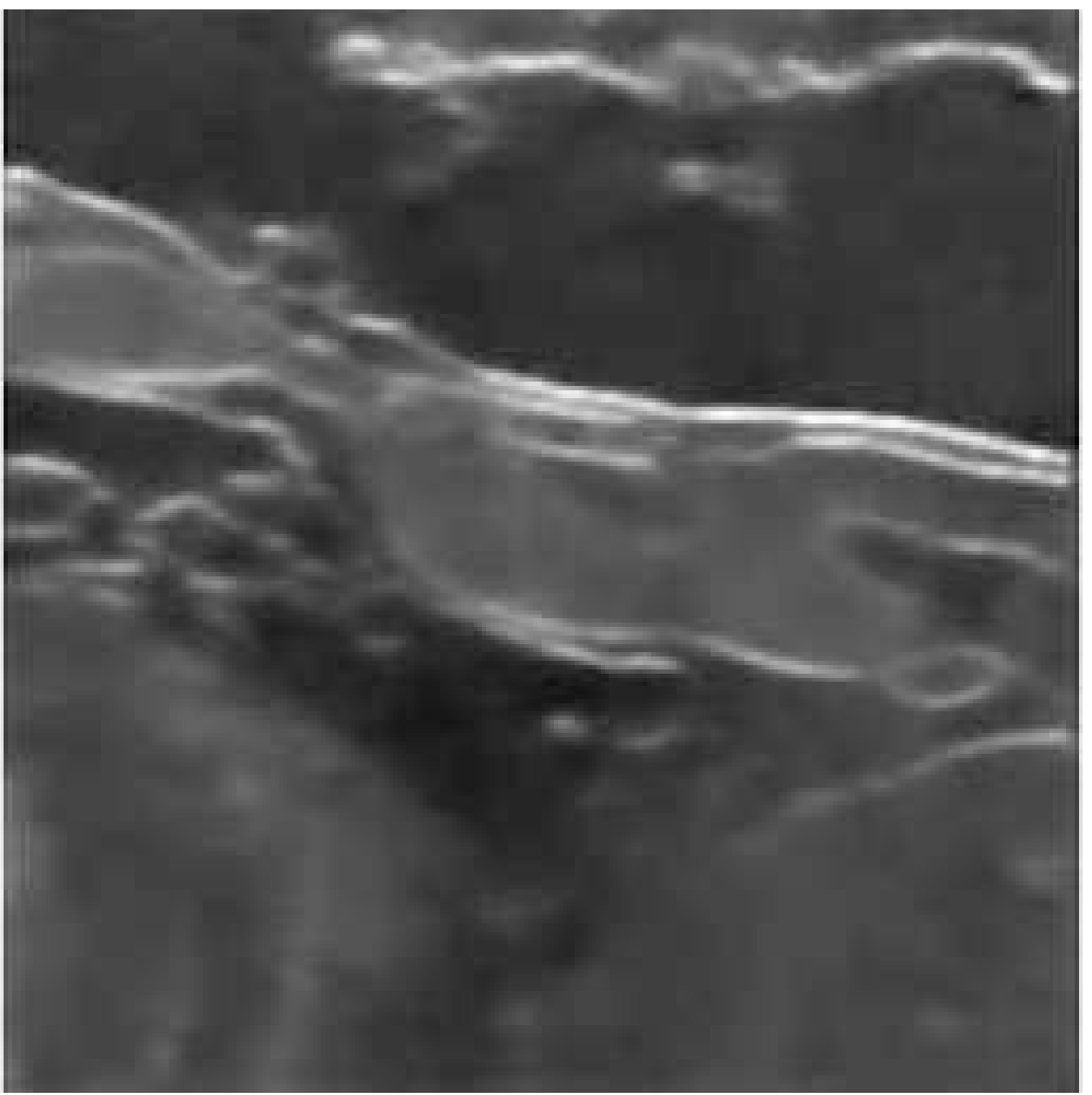} &
		\includegraphics[width = 0.199\textwidth]{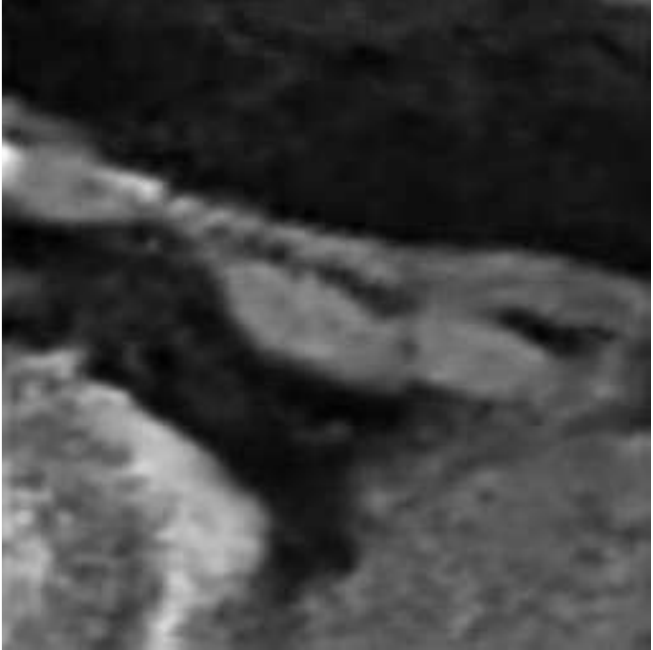}  \\ 
		&  & \small $10.85$ dB & \small $11.32$ dB & \small $19.71$ dB\vspace*{-1mm}\\
		
	\end{tabular} 
\end{minipage} 
}
	\caption{ \small \textbf{Reconstructing `CT-quality' images from ultrasound.} Depicted from left to right in each row are: real CT image used as the ground truth, corresponding simulated US image generated as described in Section \ref{ssec:data}, the output of a conventional TV despeckling algorithm, the output of our despeckling CNN-TV network, and the reconstructed CT image using our CNN-CT network. PSNR is reported relative to the ground truth CT image. }	\label{fig_large_IQ2CT}
\vspace{-0.3cm}
\end{figure}

\section{Conclusion} \label{sec:conclusions}
\vspace{-0.3cm}
In this paper, we have showed that conventional despeckling algorithms can be approximated by CNN and by that, provide a major speed-up in run-times. Moreover, we have showed that a CNN can reconstruct a CT-quality image from US IQ images within the same run-time range. With that being said, the networks were trained and tested on pairs of CT images and the corresponding simulated ultrasound images. However, in practice, the CT-US paired data is difficult to obtain. Our further research will concentrate on tackling this problem, possibly by developing an unsupervised/semi-supervised training framework.        

\bibliographystyle{IEEEbib}
\bibliography{strings,refs}

\end{document}